# MEGA: Self-Evolving Agent Optimization Infrastructure via Wisdom Graph

Jung Hwan Lee
paul@megacode.ai

Kyu Ho Lee
reeyan@megacode.ai

Gwang Hoon Yoo
gwanghoon@megacode.ai

**Abstract**

As coding agents increasingly handle implementation, the central challenge shifts from building individual agents to building an infrastructure that systematically improves them. Current approaches optimize agent systems without accumulating transferable knowledge, accumulate knowledge with-out compositional reasoning over it, and lack a mechanism for that knowledge to self-evolve through operational evidence. MEGA (Meta Evaluation-Grounded Adaptation) addresses these gaps as a self-evolving infrastructure: each optimization cycle produces durable assets, compositional reasoning over those assets guides subsequent optimization, and operational evidence refines both the accumulated wisdom and the reasoning that governs it. Layer 1 distills reusable wisdom from agent sessions through behavioral-pattern clustering and empirical A/B validation, transforming each process into a durable asset. Layer 2 decomposes these assets into atomic PCR (Primary-Context-Resultant) units within a typed Wisdom Graph and performs deductive, abductive, and inductive reasoning to expand implicit re-lations; it then assembles context-specific execution plans through compositional retrieval that surfaces bridging knowledge unreachable by embedding similarity alone. Layer 3 performs multi-agent col-laborative optimization over heterogeneous agent workflows (code nodes, LLM calls, and tool-using agents), attributing improvement effects to specific strategy changes through controlled evaluation that eliminates data variance. Evidence fed back from Layer 3 drives the self-evolution of both the curation strategies that govern wisdom composition and the optimization trajectories accumulated across runs. The result is an infrastructure in which optimizing an agent system and evolving the knowledge that guides optimization are one and the same process.

## 1 The Infrastructure Bottleneck in Agent Development

Agent-based software development is entering a new paradigm. Coding agents are increasingly performing implementation, while the human role shifts toward defining optimization objectives, designing evaluation data, and interpreting results. In this transition, the central question is no longer whether an agent can solve a single task well, but whether there exists an *infrastructure* that systematically and repeatedly improves agent systems based on data.

Tools for automating agent development are proliferating. Stateless automation loops such as the Ralph Loop [11] repeat execution cycles, and test-driven development provides verification at small scale. Prompt optimizers (DSPy [13], GEPA [2]) and workflow optimizers (AFlow [38], Flow [6]) automatically improve agent systems. Skill libraries (Voyager [31], Memento-Skills [41]) store reusable procedures for recurring tasks, while adaptive memories (Reflexion [28], ACE [39], ReasoningBank [22]) distill reasoning strategies and reflective insights from past attempts to guide future behavior. Each of these advances addresses an important aspect of the problem. What remains open is an infrastructure that, beyond merely connecting these approaches, unifies them through a self-evolving cycle in which agent sessions are distilled into verified knowledge, compositional reasoning over that knowledge guides optimization, and optimization evidence refines both the knowledge and the reasoning that governs it.

### 1.1 Three Limitations of Current Agent Development Automation

**First, optimization does not accumulate knowledge, and memory alone does not solve it.** Stateless loops [11] automate execution without learning from prior cycles [12], and test-driven verification is of-ten insufficient to guarantee correctness [36]. Prompt and workflow optimizers (GEPA [2], DSPy [13], AFlow [38], Flow [6]) can substantially improve agent systems within a single project, but the rationale behind successful optimization—which strategies worked under which conditions—remains unstructured and non-transferable. When a new project begins, strategy search restarts from scratch. Adding a memory layer does not resolve this, because the problem is not the absence of storage but the absence of compositional reasoning over what is stored.

**Second, stored knowledge lacks compositional reasoning.** Skill libraries (Voyager [31], Memento-Skills [41]) store procedural knowledge but cannot reason about *which skills to compose, in what order, under what conditions*. Skill orchestration frameworks such as AgentSkillOS [15] and SkillNet [16] have begun to address this through hierarchical organization and typed relation graphs, demonstrating that structured composition outperforms flat invocation. However, the unit of composition remains the whole skill—neither system decomposes skills into atomic sub-units, nor performs logical inference to discover implicit relations—and knowledge graphs for factual domains [4, 8, 9] face an analogous limitation. Press et al. [26] formalize this as the *compositionality gap*: models solve sub-problems yet fail to compose them, and this gap does not shrink with scale.

**Third, knowledge that is never verified through execution cannot improve.** Even if compositional reasoning were achieved, knowledge that is not tested against operational outcomes remains ungrounded. Without a feedback loop in which optimization evidence verifies whether a composition was actually effective, unverified and outdated knowledge accumulates alongside valid knowledge, and retrieval quality gradually degrades. The missing piece is a self-evolving cycle: optimization produces evidence, evidence refines accumulated knowledge, and refined knowledge guides subsequent optimization more effectively.

In summary, current agent development automation optimizes without accumulating knowledge, accumulates knowledge without compositional reasoning, and lacks a mechanism for knowledge to self-evolve through operational evidence. These three capabilities have never been unified into a single infrastructure.

## 1.2 MEGA: Evaluation-Driven Agent Development with Self-Evolving Curation

MEGA is an agent development infrastructure that transforms each optimization process into a durable asset, reasons over accumulated assets to compose context-specific execution plans, and evolves the composition logic itself through operational evidence. The key mechanism is a typed Wisdom Graph in which wisdom is decomposed into atomic PCR (Primary-Context-Resultant) units; logical reasoning discovers implicit relations among them; and compositional retrieval assembles plans that include bridging knowledge unreachable by embedding similarity alone. This reasoning substrate is not static—it self-evolves through attributed evidence produced by a multi-agent optimization loop.

1. **Layer 1** (Session-based Wisdom Generation): distills reusable wisdom from agent sessions through two-stage clustering and multi-stage quality gating including behavioral A/B validation, transforming each process into structured, reasoning-ready material.
2. **Layer 2** (Wisdom Reasoning and Curation): decomposes wisdom into atomic PCR triplets, performs deductive, abductive, and inductive reasoning to expand implicit relations, and assembles context-specific execution plans through compositional retrieval based on the Prize-Collecting Steiner Tree (PCST) [5] formulation, with ROI-driven self-evolution.
3. **Layer 3** (Evaluation-Driven Optimization): performs multi-agent collaborative optimization, attributes improvement effects to specific strategy changes through controlled evaluation, and feeds evidence back to Layer 2—driving the self-evolution of both curation strategies and optimization trajectories.

The architecture is designed so that, as optimization cycles accumulate across projects, the Wisdom Graph matures toward delivering effective optimization guidance from the first query alone.

**Scope of Disclosure.** This report presents the architectural principles and design rationale of MEGA. As portions of the system constitute proprietary technology, algorithmic details are selectively disclosed at the level of design objectives and formal properties. The content is intended to enable independent architectural evaluation and to establish the scientific foundations of the approach.

# 2 Related Work

Recent literature has rapidly advanced along three axes relevant to agent development automation: agent memory and procedural reuse, structured knowledge and skill composition, and automatic optimization of agent systems. Each axis has achieved meaningful progress in isolation; this section reviews them in turn and identifies the gaps that remain when they operate independently.

## 2.1 Agent Memory and Skill Reuse: Progress and Remaining Bottlenecks

Recent work has rapidly expanded persistent memory from simple storage toward reflective organization, procedural reuse, and cross-agent transfer. In the skill-reuse line, Voyager [31] built a library in which agents generate and accumulate code-based skills in a Minecraft environment. Memento-Skills [41] generalized this idea to general-purpose agents, automatically generating markdown skill files on the SRDP [32] framework, rewriting skills upon failure, and selecting them via an InfoNCE and offline-RL behavioral router. ProcMEM [20] formalizes reusable procedural Skills as natural-language units with activation, execution, and termination conditions, converting episodic experience into executable procedural memory.

Reflective and adaptive memory is also active. Reflexion [28] proposed verbal reinforcement learning that feeds linguistic self-reflection into subsequent attempts. Self-Refine [19] demonstrated an iterative critique-and-improve loop driven entirely by the model's own feedback. MemGPT [23] formalized long-term memory architecture through tiered memory and virtual context management. Dynamic Cheatsheet [30] builds self-curated memory at test time to adapt problem-solving strategies; ACE [39] addresses its context-collapse problem—where monolithic rewriting progressively loses accumulated knowledge—by decomposing memory into atomic bullet units with incremental delta updates (ICLR 2026).

A growing body of work extracts reusable experience from trajectories and transfers it across tasks or agents. DS-Agent [7] reuses expert cases through a case-based reasoning pipeline (ICML 2024). Agent KB [3] proposes a shared memory infrastructure across heterogeneous agent frameworks, foregrounding the foundation for collective agent intelligence. Agent Workflow Memory [33] extracts multi-step routines from successful trajectories for reuse. SWE-Exp [35] accumulates both successful and failed trajectories as a multi-layered experience bank for reusing repair expertise in software issue resolution. ReasoningBank [22] distills generalizable reasoning strategies from both successful and failed trajectories into structured memory items, demonstrating agent self-evolution at test time (ICLR 2026).

This line of work has made meaningful progress in agent memory. However, most systems treat accumulated artifacts as independent units retrieved by similarity, without modeling the relations among them—prerequisite dependencies, mutual exclusions, or conditional applicability. Whether the unit is a procedural skill, a reflective strategy, or a cross-agent experience bank, stored items remain structurally isolated: the system knows *what* it has, but not how those items relate to one another. The authors of ReasoningBank explicitly identify compositional memory as an open direction [22]. The next section examines work that imposes such relational structure.

## 2.2 Structured Knowledge, Skill Graphs, and the Limits of Current Composition

A parallel line of work imposes *structure* on knowledge—ranging from commonsense facts to operational skills—and attempts reasoning or composition over it.

ConceptNet [29] and ATOMIC [27] organize commonsense relations in fixed-type node ontologies, while Pearl [25] formalizes the notions of sufficiency and necessity in causal inference, providing theoretical tools for reasoning beyond mere correlation. On the retrieval side, GraphRAG [4] improved answer quality through entity graphs and community-level summaries, LightRAG [8] introduced graph-aware indexing with incremental updates, and HippoRAG 2 [9] unified factual, sense-making, and associative memory through non-parametric continual learning (ICML 2025). These systems demonstrate that imposing structure on *factual* knowledge enables better retrieval, but they do not address the composition of *operational* skills and strategies.

More recent work moves beyond storing skills toward orchestrating and curating them. AgentSkillOS [15] organizes skills into a hierarchical capability tree and composes them through task-specific DAG orchestration with three composition strategies (quality-first, efficiency-first, simplicity-first), demonstrating that structured composition substantially outperforms flat invocation. SkillNet [16] connects skills through a typed multi-relational graph with four relation categories (`similar_to`, `belong_to`, `compose_with`, `depend_on`) and provides a multi-dimensional quality evaluation framework, achieving consistent reward improvements across multiple agent architectures. Together, these systems validate a key premise: the bottleneck in skill-based agents is not skill availability but orchestration—how skills are organized, selected, and composed for a given task.

These systems validate that the bottleneck lies in orchestration rather than availability. However, composition still operates at the whole-skill level—neither system decomposes a compound skill into atomic sub-units whose internal dependencies can be individually reasoned about. Relational structures also remain relatively coarse: a hierarchical tree without cross-branch edges, or a small set of discrete relation types that do not distinguish the strength of association. Nor do these systems perform logical inference to discover relations that were never explicitly recorded, or self-refine through operational evidence—once a skill is registered or a relation labeled, it remains static regardless of whether downstream execution confirms or contradicts it. The next section examines whether optimization systems address these gaps from the other direction.

## 2.3 Automatic Optimization of Agent Systems: From Prompts to Workflows

Research on automatically improving agent system performance is growing rapidly and bifurcating into two tiers defined by the optimization object.

The first tier optimizes prompts, instructions, demonstrations, and related settings over fixed LM programs. DSPy [13] frames LM calls as declarative programs whose prompts and examples are automatically optimized by a compiler. MIPROv2 [21] jointly searches instructions and few-shot demonstrations for multi-stage LM programs, demonstrating strong improvements within fixed program structures. GEPA [2] reflectively analyzes trajectories—including tool calls and intermediate outputs—to evolve prompt updates, achieving strong results across multiple benchmarks for prompt- and instruction-level optimization. Feedback Descent [14] generalizes text-level optimization through pairwise comparison with cumulative textual rationale. Optimas [34] optimizes heterogeneous configurations including prompts, hyperparameters, and model parameters while maintaining component-level local rewards for globally aligned optimization of compound AI systems. The common thread is that, even as the optimization surface broadens, the primary intervention target remains internal to a fixed system—prompts, instructions, demonstrations, hyperparameters, and component settings—rather than code-level workflow modification itself.

The second tier treats the workflow as a mutable optimization target at the code-node level, rather than refining prompts within an LM program. AFlow [38] formalizes agentic workflow optimization as a search problem over code-represented workflows; the core of optimization lies not in better-phrasing prompt chains but in restructuring LLM-invoking nodes, their connections, task decomposition, execution order, and control flow at the code level. Flow [6] represents workflows as activity-on-vertex graphs and performs dynamic subtask allocation and continuous workflow refinement based on historical performance and prior AOV structures, focusing on restructuring execution procedures—parallelism, dependencies, and module boundaries—rather than refining individual prompt wording. The defining characteristic of this tier is that it treats agent systems as mutable code/graphs and optimizes node composition, edge structure, task decomposition, and execution policies as first-class optimization variables.

The two tiers share two structural limitations. First, neither accumulates *why* a particular strategy worked under particular conditions as structured, cross-project transferable knowledge. The rationale behind successful optimization typically remains in execution logs or search traces rather than being orga-nized as transferable optimization knowledge that becomes the starting point for the next project. Consequently, even as powerful automatic optimization becomes available, the strategy search restarts from scratch when a new system or new task is encountered. Second, both tiers assume that evaluation data is externally provided and treat it as fixed input, even though a growing body of work demonstrates that synthetic data meaningfully improves agent capabilities—SWE-Next [18] for coding agents, ProgSearch [24] for web research agents, and SandMLE [42] for ML engineering agents—leaving the evaluation dataset itself outside the optimization loop.

Recent empirical studies independently corroborate the first limitation in the broader automation landscape. Yu et al. [36] demonstrate that test-driven verification is often insufficient—345 erroneous patches passed SWE-bench tests due to inadequate coverage—and Joshi et al. [12] characterize existing SWE-bench agents as stateless, treating each issue independently without cross-task experience accumulation. Stateless automation loops such as the Ralph Loop [11] exhibit the same pattern: each cycle does not learn from prior cycles, so productive work and waste are automated equally. Taken together, these observations suggest that current automation tools do not structure *which optimization approach was effective for which type of problem under which conditions* as transferable knowledge.

The preceding review reveals a consistent pattern. Memory systems accumulate experience but do not decompose it into compositionally reasonable units (Section 2.1). Graph-based systems and skill orchestration frameworks impose structure but do not decompose skills into atomic reasoning units, discover implicit relations through logical inference, or self-refine through execution evidence (Section 2.2). Optimization systems improve agent performance but do not structure transferable knowledge about why a strategy succeeded, nor do they generate the evaluation data on which optimization depends (Section 2.3). MEGA addresses these gaps not by combining existing tools but by introducing a new architectural element—a typed Wisdom Graph with logical reasoning and evidence-driven self-evolution—that transforms the relationship between optimization and accumulated knowledge.

# 3 Architecture and Design Principles

MEGA's core function is evaluation-driven optimization of agent systems. The three layers form a cycle in which optimization produces evidence, evidence refines the Wisdom Graph, and the refined graph guides subsequent optimization:

$$\mathcal{W}^{(t)} = \mathcal{L}_1\big(\mathcal{T}^{(t)} \cup \mathcal{T}_3^{(t-1)}\big), \quad \Pi^{(t)} = \mathcal{L}_2\big(\mathcal{W}^{(t)}, \mathcal{E}^{(t-1)}\big), \quad \mathcal{E}^{(t)}, \mathcal{T}_3^{(t)} = \mathcal{L}_3\big(\Pi^{(t)}\big) \tag{1}$$

Here, $t$ indexes the outer self-evolution cycle of MEGA (distinct from the Seed-Epoch index $e$ used inside Layer 3, defined in Section 6.4). $\mathcal{T}^{(t)}$ denotes general agent session traces, while $\mathcal{T}_3^{(t)}$ denotes execution sessions produced by Layer 3's optimization loop. The union $\mathcal{T}^{(t)} \cup \mathcal{T}_3^{(t-1)}$ indicates that Layer 1 consumes both trace sources through a single PCR extraction pipeline without distinguishing their origin. We initialize $\mathcal{E}^{(0)} = \varnothing$ and $\mathcal{T}_3^{(0)} = \varnothing$, so the first cycle operates without prior evidence.

The three layers form a self-reinforcing cycle. Layer 3 produces two outputs that flow backward: evaluation evidence $\mathcal{E}^{(t)}$ refines Layer 2's curation and planning, while execution sessions $\mathcal{T}_3^{(t)}$ become new input to Layer 1 for fresh wisdom distillation. This structure ensures that optimization outputs reshape the knowledge that guides future optimization. Figure 1 visualizes the cycle.

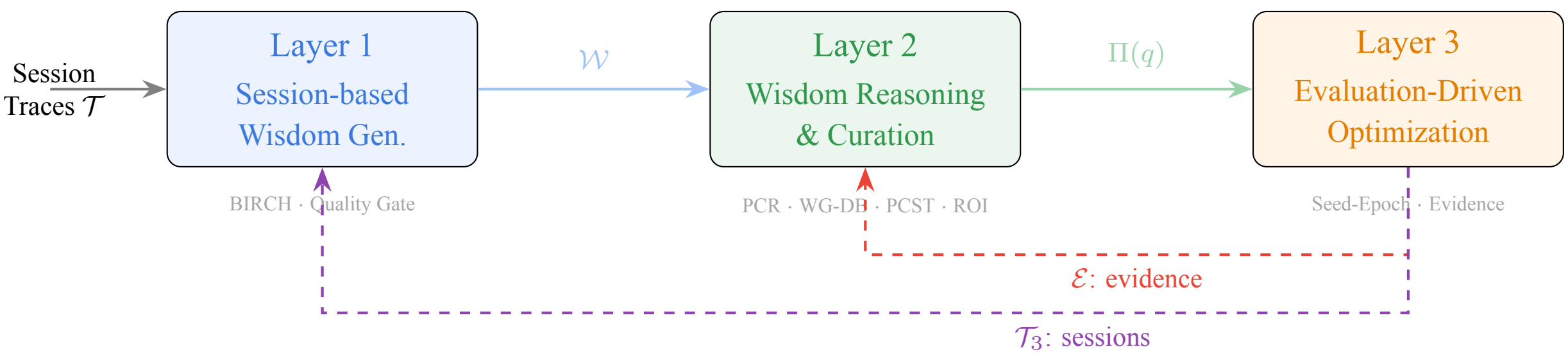

Figure 1: MEGA architecture. $\mathcal{W}$ comprises four wisdom types (skill, strategy, curation pattern, optimization trajectory). $\Pi(q)$ is a context-specific execution plan. Evidence $\mathcal{E}$ (red) includes verdicts, curation feedback, and optimization trajectories that Layer 2 uses to generate curation pattern and optimization trajectory wisdom within WG-DB. Execution sessions $\mathcal{T}_3$ (purple) feed into Layer 1 as new extraction input.

## 3.1 Inter-Layer Artifact Contract

Specifying each layer's inputs and outputs makes the system boundaries precise. Table 1 summarizes this contract.

Table 1: Inter-layer artifact contract. The input, output, and core transformation of each layer are specified.

| Layer | Input | Output | Core Transformation |
|---|---|---|---|
| 1 | Session traces $\mathcal{T} \cup \mathcal{T}_3$ | Wisdom assets $\mathcal{W}$ | Clustering + quality gating |
| 2 | $\mathcal{W}$ + evidence $\mathcal{E}$ | Execution plan $\Pi(q)$ | PCR reasoning + PCST + ROI |
| 3 | $\Pi(q)$ + execution results | Evidence $\mathcal{E}$, sessions $\mathcal{T}_3$ | Fixed-seed attribution |

## 3.2 Four Types of Wisdom

Four distinct types of wisdom coexist within the system, each carrying causal judgment:

Table 2: Four types of wisdom. MEGA accumulates not only skills but also strategies, curation patterns, and optimization trajectories as causal judgments.

| Type | Description | Producer | Consumer |
|---|---|---|---|
| Skill | Reusable procedural knowledge | Layer 1 | Layer 2 retrieval |
| Strategy | Decision rules and judgment criteria | Layer 1 | Layer 2 curation |
| Curation pattern | Which skills and strategies worked in what order and under what conditions | Layer 2/3 | Layer 2 planning |
| Optimization trajectory | Which optimization trajectories succeeded or failed | Layer 3 | Layer 2/3 |

This distinction matters because MEGA's unit of learning extends beyond "what skills exist" to encompass "which combinations worked in which contexts," "which execution orders were reproducible," and "which trajectories reduced failures."

Curation patterns, in particular, create substantial value in practice. Many failures stem not from the absence of individual skills but from invoking the right skills in the wrong order, under the wrong conditions, or with premature confidence. Knowing which orderings were stable, which evidence thresholds must be met before proceeding to the next step, and which failure signals should trigger a return to exploration mode—these are Wisdom-level judgments that cannot be captured at the Knowledge level.

Optimization trajectories are likewise not mere logs. They encode operational wisdom about "what changes were attempted, what improved, and what broke," serving as higher-order wisdom that Layer 2 references when determining exploration/exploitation strategies for future tasks. All four types of wisdom share PCR semantics within WG-DB and are subject to the same reasoning and curation mechanisms.

### 3.3 Wisdom Graph as a Reasoning Substrate

In the DIKW pyramid of information science [1], the transition from Knowledge to Wisdom has traditionally been discussed in terms of value judgment and effectiveness. We reinterpret that transition computationally.

A Knowledge Graph is a network representation of information. It structures factual relations as nodes and edges—"A is related to B," "X is a type of Y." ConceptNet [29] can store "a car runs on a road," but it cannot *derive* on its own that "when the road is wet, the car's braking distance increases." This is structured *storage*, not *reasoning*. Existing memory systems and knowledge graphs operate at this level.

MEGA's WG-DB operates beyond this representational level. We use the term *wisdom* to denote a reusable causal judgment of the form "under context $v_C$, action $v_P$ produces resultant $v_R$, realized via method $m$"; each wisdom asset decomposes into atomic PCR (Primary-Context-Resultant) triplets with typed dependencies—prerequisite, exclusion, conditional branching, and fallback—among sub-units, and four disjoint subtypes (skills, strategies, curation patterns, optimization trajectories; Table 2) share these PCR semantics. The *Wisdom Graph* (WG-DB) is then the directed typed multi-graph $\mathcal{G} = (V, E)$ whose nodes are these PCR components and whose edges carry Sufficiency/Necessity scores $\sigma(e) = (S(e), N(e)) \in [0, 1]^2$ inspired by Pearl's [25] causal calculus, with a role-fluid node pool that lets the same concept serve as Primary in one triplet and as Context in another. Over this structure, MEGA performs deduction (transitive logical inference), abduction (reverse logical inference), and induction (statistical pattern discovery) to derive relations that were never explicitly recorded, and operational evidence feeds back to enable the graph to self-refine. Section 5.2 formalizes the triplet, the scoring, and the inference procedures in full.

Furthermore, WG-DB does not remain a reasoning substrate for a single project. Skills, strategies, curation patterns, and optimization trajectories accumulated across multiple projects mitigate the cold-start problem of new projects, enabling them to begin from previously validated structures. This is what makes WG-DB a collective intelligence substrate transferable across projects.

## 4 Layer 1: Session-based Wisdom Generation

Layer 1 distills raw conversation sessions into reusable wisdom assets. This is not simple log storage but knowledge distillation: it extracts structured rules, workflows, and strategies from sessions spanning hundreds to thousands of turns, and passes only those that survive multi-stage quality verification—structural filtering, self-evaluation scoring, and behavioral A/B testing—to Layer 2. As a prerequisite, privacy-preserving filtering—including secret masking and path anonymization—is applied client-side before any data leaves the local environment, ensuring that the entire pipeline operates on sanitized traces. Figure 2 summarizes this pipeline end-to-end.

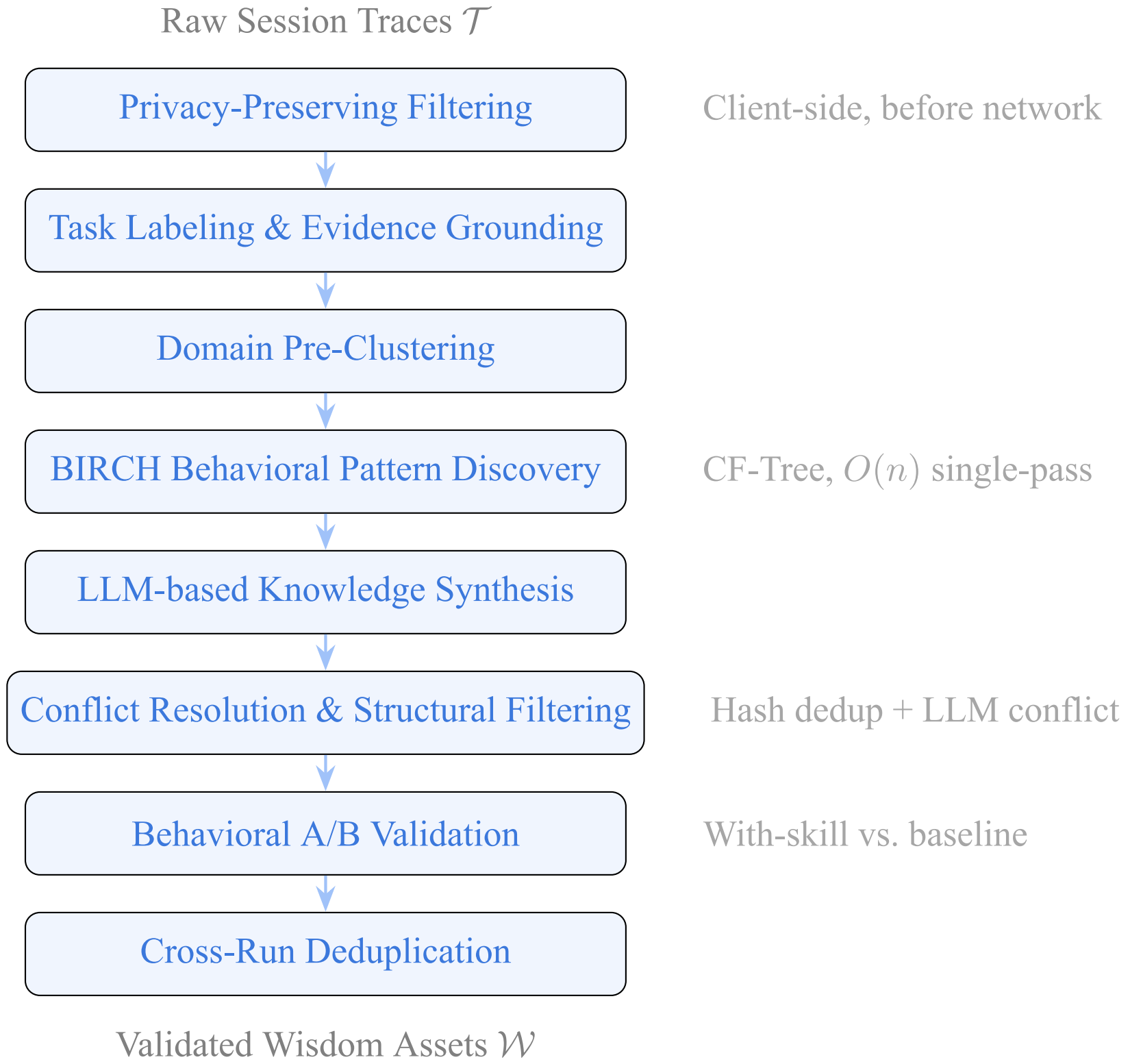

Figure 2: Layer 1 extraction pipeline. Client-side privacy filtering precedes all processing. BIRCH discovers behavioral patterns in a single $O(n)$ pass, and multi-stage quality gating admits only verified wisdom.

### 4.1 Distillation over Raw Trace Storage

Many memory systems [23, 28, 33] either feed entire sessions back as long context or vectorize session fragments for retrieval in similar situations. This approach is useful for short FAQ-style problems but reveals three limitations in complex agent workflows. First, raw traces mix signal and noise, making the unit of reuse unclear. Second, when sessions that solved the same problem with different strategies are mixed, retrieval may return conflicting instructions. Third, returning entire sessions blurs the boundaries of privacy filtering and evidence attribution.

Distilled wisdom, by contrast, first determines "what is the essential pattern" before storing it. Layer 1's task is not text compression but the separation of only those structures from a session that have reuse value. Only by explicitly structuring which error signal was the starting point, which action was the key fix, which result was verified, and which exception conditions apply can Layer 2 accept the output as a reasoning substrate. In other words, Layer 1's distillation is not storage optimization but formal preprocessing required for Layers 2 and 3 to function.

This difference manifests clearly in operational quality. Raw trace memory shows "similar past conversations," whereas distilled wisdom provides "candidate actions to execute now and their conditions." While raw traces may assist human practitioners, they still impose substantial interpretation cost on machines; distilled wisdom is an artifact directly composable by the next layer. MEGA therefore redefines long-term memory not as a session-preservation problem but as a transformation-to-actionable-form problem.

### 4.2 Two-Stage Clustering: Domain Separation and Behavioral Pattern Discovery

The technical core of Layer 1 is two-stage clustering.

**Stage 1 (Domain Pre-Clustering):** Sessions are classified by domain via greedy assignment based on embedding similarity, where domains emerge from the data rather than being predefined. This stage narrows the search space for subsequent BIRCH clustering, improving efficiency.

**Stage 2 (Behavioral Pattern Discovery):** BIRCH [40] is applied within each domain cluster. Each sub-cluster is represented by a Clustering Feature (CF) tuple $\text{CF} = (N, \vec{LS}, SS)$, where $N$ is the count, $\vec{LS} = \sum_{i=1}^{N} \vec{x}_i$ is the linear sum, and $SS = \sum_{i=1}^{N} \|\vec{x}_i\|^2$ is the squared sum. CFs are additive—merging two disjoint sub-clusters requires only:

$$\text{CF}_1 + \text{CF}_2 = \left(N_1 + N_2,\ \vec{LS}_1 + \vec{LS}_2,\ SS_1 + SS_2\right) \tag{2}$$

This $O(1)$ merge operation is what enables single-pass $O(n)$ processing over the full session set. Using the radius-based threshold criterion, a new session embedding $\vec{x}$ is assigned to the closest existing sub-cluster when the resulting radius remains within the threshold $T$:

$$R(\text{CF}') = \sqrt{\frac{SS'}{N'} - \frac{\|\vec{LS}'\|^2}{N'^2}} \leq T \tag{3}$$

where $\text{CF}' = \text{CF} + (1, \vec{x}, \|\vec{x}\|^2)$. If the condition is violated, a new sub-cluster is created. This mechanism produces an adaptive number of sub-clusters without requiring a pre-specified $k$—a critical property since the number of behavioral patterns in agent sessions cannot be known a priori. The CF-Tree structure further ensures memory efficiency by bounding tree width through a branching factor.

This two-stage structure transforms Layer 1 from "conversation storage" into behavioral-pattern-based knowledge distillation. Sessions that are semantically similar but behaviorally distinct—sessions that solved the same error with different strategies—are separated and synthesized into independent wisdom assets.

### 4.3 Synthesis and Structural Filtering

Behavioral patterns extracted from clusters are synthesized by an LLM into structured wisdom. This synthesis is not simple summarization but a process of generalizing common patterns within a cluster and specifying exception conditions.

Synthesized candidates undergo structural filtering: (1) hash-based exact deduplication ($O(n)$), (2) LLM-based conflict detection—identifying and resolving contradictory instructions and mutually exclusive candidates, (3) cross-run embedding-similarity deduplication—merging candidates extracted from different sessions that are semantically identical. This structural filtering prevents knowledge-base contamination.

For sessions where extraction could not complete (due to network failures, timeouts, etc.), a pending management mechanism preserves partial results, enabling subsequent retries to skip already-completed stages and resume processing.

### 4.4 Behavioral A/B Validation

Evaluating extracted knowledge typically relies on proxy signals—metadata quality, LLM-generated ratings, or structural completeness. These signals assess the artifact itself rather than its effect on downstream agent behavior. MEGA Layer 1 introduces behavioral A/B validation, which directly measures the causal impact of each wisdom candidate on task performance. Test cases are generated alongside each wisdom candidate using an automated teacher-student evaluation protocol [10]: a high-capability model produces both the wisdom document and its associated test suite; the same or a stronger model then executes the test suite under treatment and control conditions. For each wisdom candidate that passes synthesis and structural filtering, these test cases are executed in parallel under two conditions:

- **Treatment group**: execution with the extracted wisdom injected into the system prompt
- **Control group**: execution with the same input under the baseline condition without wisdom

The outputs of the two runs are compared to empirically measure performance lift (accuracy improvement) and token efficiency (cost change). These two ROI axes form the basis for accept/reject decisions.

Three key differentiators define this design. First, the evaluation criterion is observable behavioral change, not subjective quality judgment. Second, the two axes of performance lift and token efficiency are independently valuable—wisdom that significantly reduces cost even at equivalent accuracy can still pass.

Third, this validation runs automatically for all candidates, maintaining wisdom quality at the operational stage without human review.

Consequently, Layer 1's quality assurance comprises three tiers: structural filtering (deduplication and conflict removal), self-evaluation scoring, and behavioral A/B validation (empirical ROI measurement). Self-evaluation scoring collects factual, countable properties of the generated document—such as the ratio of rules backed by concrete commands or code examples—and aggregates them via a weighted harmonic mean that naturally penalizes any single weak dimension. To prevent self-inflation, the system cross-verifies the model's self-reported counts against an independent parse of the output, adopting the independently parsed values—either directly or as an upper bound—for each objective dimension. Only wisdom that passes this entire pipeline enters Layer 2's reasoning substrate.

# 5 Layer 2: Wisdom Reasoning and Curation

Layer 2 connects wisdom distillation (Layer 1) with optimization (Layer 3). Before any asset enters the graph, a security screening gate inspects the full skill package—instruction documents, prompt templates, and attached scripts—for two classes of contamination: instruction-level threats such as prompt injection or workflow hijacking embedded in text, and execution-level threats such as exfiltration logic or destructive commands in code. Only packages that pass screening become eligible for graph insertion. At build-time, Layer 2 decomposes incoming wisdom into atomic PCR units, normalizes nodes, and expands the graph through logical inference; at query-time, it assembles context-specific execution plans through compositional retrieval and evidence-driven curation.

## 5.1 Compositional Retrieval beyond Skill Selection

On the surface, MEGA may appear similar to a skill combination system in that it bundles multiple skills into a plan. However, the essence of a combination engine is scoring and combining a given candidate pool. In that case, the graph is typically close to a static index, and the retrieval result reduces to selecting a subset of existing items. In MEGA, by contrast, Layer 2 *extends* the graph through reasoning before combination. That is, the system does not stop at selecting already-stored items; it infers bridge, prerequisite, exclusion, and fallback relations among stored items that have not yet been made explicit.

This distinction becomes sharper at query time. A combination engine typically selects "three similar skills" and lists them in a prompt. MEGA, however, first identifies which types of wisdom are jointly needed, then decomposes the role each wisdom plays within the plan. Some items become direct execution targets, others become conditionals, and still others become fallback rules activated only upon failure. Plan generation in MEGA is therefore not the construction of a candidate set but the placement of role-differentiated artifacts into an execution graph.

Moreover, MEGA accumulates negative evidence alongside positive evidence. Even if a skill combination engine learns "combinations that worked," it cannot reduce repeated failures without structuring why a particular combination failed under particular conditions. Through trajectories and verdict feedback, MEGA produces conditional wisdom about "when *not* to invoke." This is why MEGA's graph is not a simple recommendation surface but a reasoning surface that simultaneously learns the boundaries of success and failure.

## 5.2 PCR Atomic Decomposition and the Role-Fluid Graph

Each wisdom asset undergoes hierarchical PCR decomposition: a compound skill is factored into atomic sub-skill units, each represented as a PCR triplet. The logical dependencies among sub-skills—prerequisite ordering, mutual exclusion, and conditional branching—are captured as typed edges within the PCR graph. This decomposition enables Layer 2 to reason about the internal structure of complex skills, not merely their external interfaces, thereby supporting compositional retrieval at sub-skill granularity.

**Definition 1** (Wisdom Triplet)**.** $\omega = (v_P, v_C, v_R, m) \in V^3 \times \mathcal{M}$. $v_P$ (Primary): the core action, $v_C$ (Context): the application condition, $v_R$ (Resultant): the expected outcome, $m$ (Method): implementation de-tails. Each edge $e = (u, v)$ carries a dual-axis score inspired by Pearl's [25] notions of sufficiency and necessity:

$$\sigma(e) = (S(e),\ N(e)) \in [0, 1]^2 \tag{4}$$

$S(e)$ denotes the degree to which the target node follows when the source node is present (sufficiency), and $N(e)$ denotes the degree to which the target node is unlikely to occur without the source node (necessity).

WG-DB uses a Role-Fluid node pool: the same node can serve as $P$ in one triplet and as $C$ in another. Unlike structures such as ConceptNet or ATOMIC that assign fixed types to nodes, this design allows the same concept (e.g., "Git branching strategy") to function naturally as an action to perform (Primary) in one context and as a precondition (Context) in another. This role fluidity naturally generates cross-skill connections and forms a richer relational network while reducing the node count compared to fixed-type graphs.

### 5.3 PCR Reasoning: PCR as a Logical Inference Layer

PCR triplets are not merely a storage format. They are what transforms Layer 2 from "a knowledge graph with retrieval preprocessing" into a logical inference layer. The directional relations among $P$, $C$, and $R$ provide the structural basis for three classical modes of reasoning.

**Deduction: Transitive Causal Inference.** When $P \xrightarrow{\sigma_1} C$ and $C \xrightarrow{\sigma_2} R$ relations exist in the PCR structure, $P \xrightarrow{\sigma'} R$ is inferred transitively. Multiplicative score decay is applied at each hop so that longer-range inferences receive exponentially more conservative scores, suppressing false positives. For example, if "execute index rebuild ($P$)" is linked to the condition "when ORM N+1 error occurs ($C$)," and that condition leads to "query performance improvement ($R$)," the transitive relation $P \to R$—index rebuild directly improves query performance—is inferred. This inference applies not only within a single triplet but equally to cross-wisdom causal chains formed when one wisdom's result ($R$) matches another wisdom's action ($P$).

**Abduction: Reverse Causal Inference.** When $P \to R$ and $C \to R$ edges exist—that is, when action $P$ and condition $C$ share the same result $R$—the system back-traces "why the same result?" to discover a hidden $P \to C$ relation. The key premise is that the Necessity score of $C \to R$ must be sufficiently high: there must be a strong dependency such that $R$ almost certainly requires $C$, making the abductive inference "$P$ must also go through $C$ to reach $R$" reliable. A Bayesian posterior estimate combining observed result coverage, the causal strength of $C \to R$, and semantic relatedness yields the confidence of candidate relations. This is particularly useful for discovering "preconditions that were never explicitly documented."

**Induction: Statistical Pattern Discovery.** When multiple $P$ nodes are simultaneously connected to the same $C$ and the same $R$—$P_n \to C$, $P_n \to R$—the system examines the possibility of a direct relation $C \to R$. A hypergeometric test verifies whether this co-occurrence exceeds chance, and the $C \to R$ edge is inferred only if statistically significant.

The three reasoning types are applied iteratively. First, the edge set is monotonically expanding: edges are only added, never removed, and a deduplication index prevents the same edge from being added twice. Then, each reasoning mode applies stage-specific acceptance thresholds—minimum causal strength for deduction (with a dynamic penalty that rises logarithmically with target in-degree), Bayesian posterior bounds for abduction, and hypergeometric significance tests for induction—so that progressively weaker inferences are filtered out. When a reasoning cycle produces zero new edges that survive these thresholds, the system has reached a fixed point and terminates.

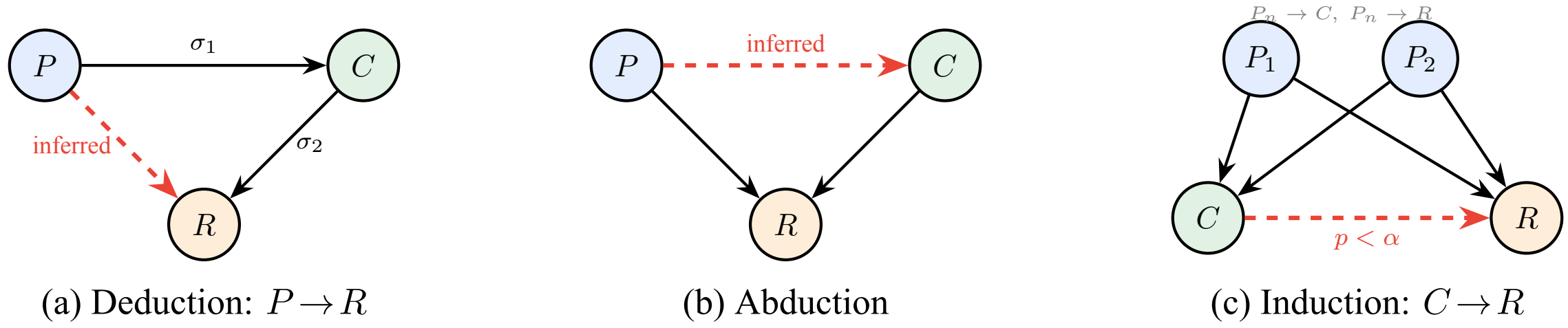

Figure 3: PCR Reasoning. The directional $P \to C \to R$ structure of PCR provides the structural basis for three classical modes of reasoning. Solid lines denote observed edges; red dashed lines denote inferred edges.

### 5.4 PCST-Based Compositional Retrieval

Embedding similarity search returns only knowledge directly similar to the query. It cannot discover "bridging knowledge"—knowledge essential for connecting multiple skills but exhibiting low similarity to the query—which is critical for complex tasks.

WG-DB formalizes this as a Prize-Collecting Steiner Tree (PCST) [5] problem over the undirected projection $\bar{\mathcal{G}}$ of WG-DB, obtained by retaining the maximum causal strength for each undirected node pair:

$$T^* = \arg \max_{\substack{T \subseteq \bar{\mathcal{G}} \\ T \text{ connected}}} \left[ \sum_{v \in V(T)} \pi(v; q) - \sum_{e \in E(T)} c(e) \right] \tag{5}$$

$\pi(v; q)$ is a prize proportional to query relevance, and $c(e)$ is a cost proportional to causal weakness. Since $c(e) \geq 0$, any optimal connected solution is a tree. Non-seed nodes in the PCST solution—those with low query similarity but essential for connecting seeds—constitute the bridging wisdom.

Crucially, PCST does not retrieve only skill-type wisdom. Since all wisdom types coexist in the same graph, PCST performs mixed-type subgraph assembly that cross-composes heterogeneous wisdom types including strategies, curation patterns, and trajectories.

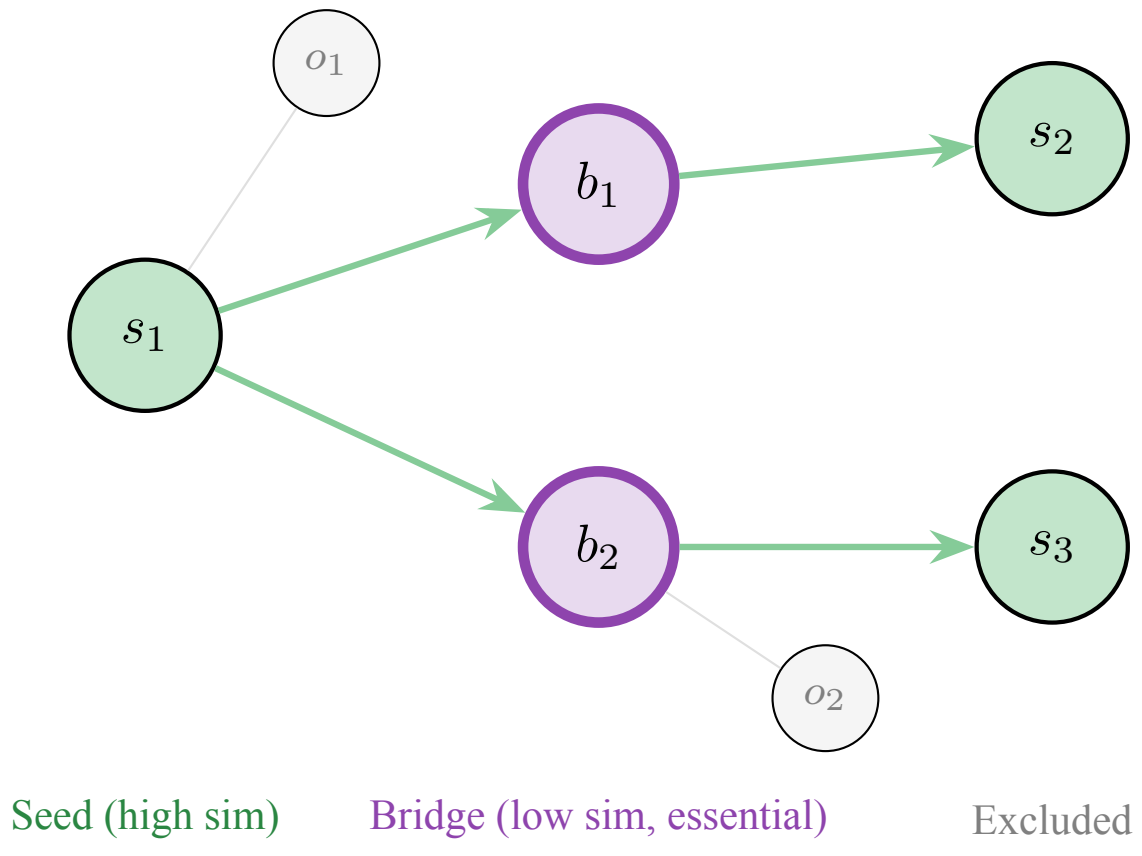

Figure 4: PCST-based retrieval. Seed nodes have high query similarity, while Bridge nodes have low similarity but are included in the Steiner tree solution as essential connectors between seeds.

### 5.5 From Subgraph to Execution

The output of PCST is not an immediately executable script. It is a reasoning subgraph representing "what should be considered together." Here, Layer 2 performs a second transformation: converting the mixed-type subgraph into a plan artifact with execution ordering. Skill nodes become execution candidates; strategy nodes become gating rules and priority constraints; curation patterns provide topological constraints and dependency ordering; and trajectories provide initial bias in conservative/exploratory branches.

This transformation is what takes MEGA one step beyond a "retrieval system." Instead of listing retrieved results in the prompt context, it separates the roles of retrieved wisdom: some are executed immediately, some are used as pre-execution conditions, and some are preserved as fallback paths upon failure. The output $\Pi(q)$ of Layer 2 is therefore not a simple top-$k$ list but the linearization of a plan graph with assigned execution rules. This distinction separates MEGA from skill combination approaches.

## 5.6 ROI-Based Wisdom Curation: Self-Evolution, Not Simple Ranking

The second core role of Layer 2 is to continuously evaluate and refine the effectiveness of accumulated wisdom. This is not simple ranking—it is a self-evolution mechanism in which the system calibrates its own judgments against historical performance. The formulations below convey the key design intuitions of this mechanism; detailed parameter settings and statistical assumptions are omitted.

**Predicted Utility Estimation.** For each wisdom $\omega_i$:

$$p_i = \tau(\text{sim}_i, \text{stage}_i) \times \eta(\omega_i) \tag{6}$$

$\tau(\text{sim}_i, \text{stage}_i)$ is the historical transfer rate—the empirically observed average effectiveness for items at a similar similarity level and workflow stage. It starts from a prior distribution at cold start and converges to the empirical mean as observations accumulate. $\eta(\omega_i)$ is a saturation-based evidence confidence that scales with feedback accumulation, guaranteeing a minimum exploration opportunity even for items with no evidence.

**Step-Level Aggregation.** $P_{\text{step}} = 1 - \exp(-\sum_i p_i)$. A saturation form is used to reflect diminishing marginal utility of redundant skills.

**Adaptive Gating.** The system adaptively switches between a cold-start gate (permissive, exploration-first) and a warm gate (conservative, LCB-based) according to system confidence:

$$g_{\text{roi}}(\omega_i) = \begin{cases} p_i \geq \theta_{\text{cold}} & \text{if conf} < \theta_c \\ \text{lcb}_i > \theta_{\text{warm}} & \text{otherwise} \end{cases} \tag{7}$$

**Four-Axis Composite Scoring.** For items that pass the gate, a composite score is computed across four axes: embedding similarity (*sim*), evidence strength (*evi*), stage fitness (*stage_fit*), and benchmark (*bench*). Missing axes are handled by renormalizing weights over the available axes.

**Portfolio ROI Estimation and Calibration.** Expected performance is estimated as a score-weighted average, and a Correction Factor $\text{CF} = \mathbb{E}[\delta_{\text{actual}}/\delta_{\text{predicted}}]$ is applied to calibrate against past prediction–outcome discrepancies. This CF constitutes the system's "self-awareness": it learns how accurate its own predictions have been and calibrates subsequent predictions accordingly.

## 5.7 Wisdom Self-Evolution: A System That Improves with Use

After each curation session completes, per-wisdom attributions are extracted from the session outcome, decomposing the session-level result into individual contribution levels, observed effects, and optional revision suggestions. These attributions drive self-evolution along three axes. Figure 5 illustrates the feedback loop.

**Individual-Level Calibration.** Per-wisdom effectiveness predictions are calibrated against observed outcomes. The system maintains a transfer function $\tau$ that maps similarity distance and workflow stage to expected effect:

$$\tau^{(n)} = \begin{cases} \tau_{\text{prior}} & n = 0 \\ \dfrac{\bar{\tau}_{\text{obs}} + \tau_{\text{prior}}}{2} & 0 < n \leq k \\ \bar{\tau}_{\text{obs}} & n > k \end{cases} \tag{8}$$

where $n$ is the observation count and $k$ is the blending threshold. Cold-start items receive prior-based estimates; well-observed items converge to their empirical effectiveness profile. This is the transfer rate $\tau(\text{sim}_i, \text{stage}_i)$ referenced in the predicted utility estimation of Section 5.6.

**Compositional-Level Reinforcement.** The system records which wisdom *combinations* succeeded or failed under similar query conditions. A combination **c** is promoted to a curation strategy when it meets a frequency–quality threshold:

$$\text{promote}(\mathbf{c}) \iff |\mathcal{S}(\mathbf{c})| \geq n_{\text{min}} \wedge \bar{r}(\mathbf{c}) \geq r_{\text{min}} \tag{9}$$

where $\mathcal{S}(\mathbf{c})$ is the set of sessions in which **c** was observed and $\bar{r}(\mathbf{c})$ is the mean rating across those sessions. Promoted patterns are reusable templates suggested as in-context guidance in subsequent sessions. Combinations that consistently co-occur with high ratings are further promoted to orchestration strategies that inform both plan decomposition and candidate prioritization. Per-wisdom reviews, maintained as compact summaries updated through delta operations, provide qualitative context that sharpens stage-fitness judgments across cycles.

**Content-Level Evolution.** When revision suggestions accumulate beyond a threshold for a given wisdom item, the item is flagged as an evolution candidate:

$$\text{evolve}(\omega) \iff |\mathcal{U}(\omega)| \geq \theta_{\text{evo}} \tag{10}$$

where $\mathcal{U}(\omega)$ is the set of unconsumed revision suggestions for wisdom $\omega$ and $\theta_{\text{evo}}$ is the evolution threshold. The accumulated suggestions are preserved as durable signal for the maintenance pipeline (Section 5.8), which makes the final update decision with full context.

**Offline Pattern Consolidation.** When a pattern group accumulates sufficient members, an offline synthesis step—executed in batch, outside the live curation loop—compresses them into a single best-practice pattern capturing the common successful structure and weighted average performance. Previous syntheses serve as context for incremental refinement, preventing pattern proliferation while preserving distilled operational knowledge.

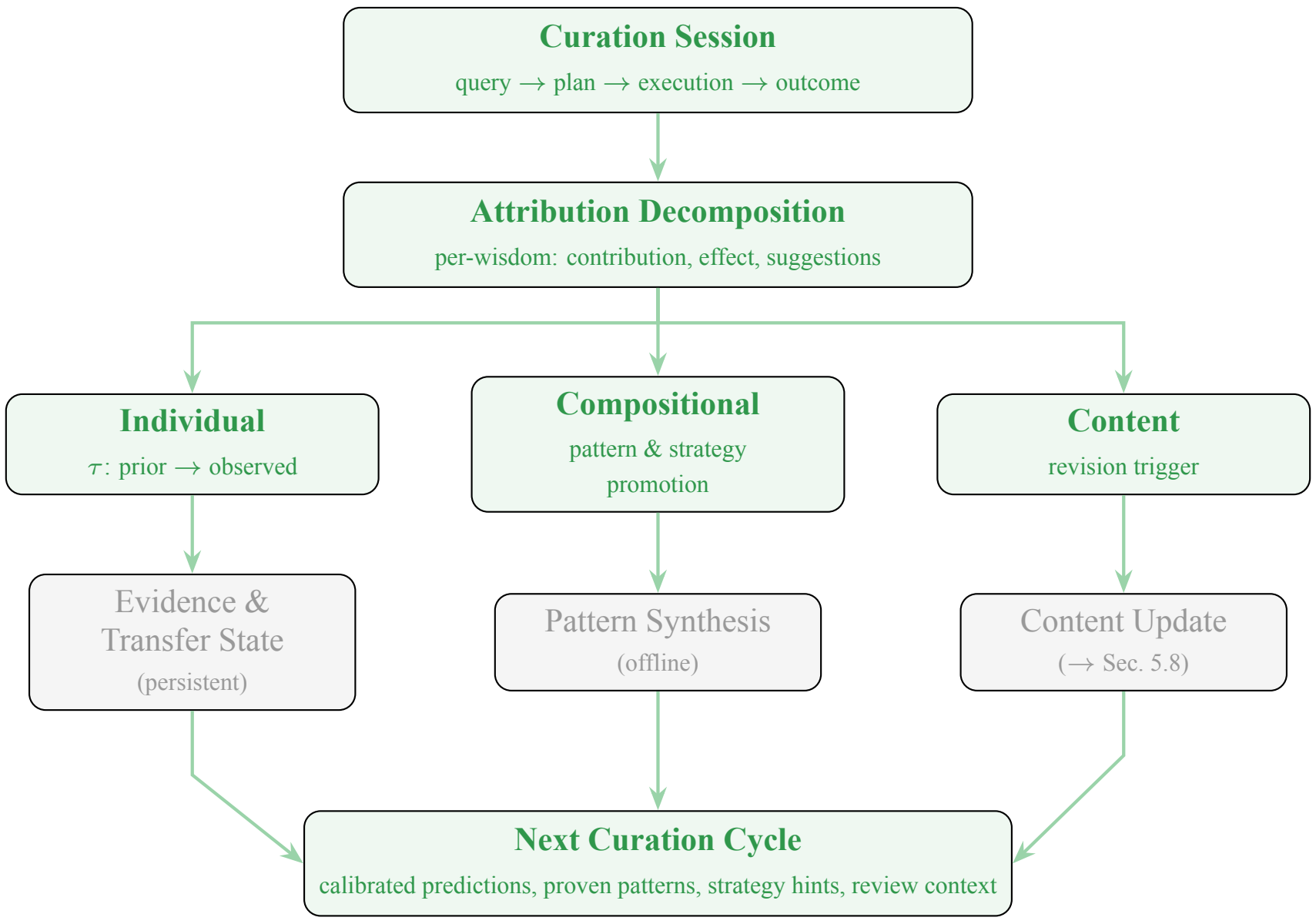

Figure 5: Self-evolution feedback loop. Per-wisdom attributions from each curation session update three axes—individual effectiveness calibration, compositional pattern and strategy promotion, and content evolution triggers. Offline synthesis consolidates patterns into best-practice templates. All learned state feeds into the next curation cycle.

### 5.8 Wisdom Maintenance: Auto Graph Hygiene

Self-evolution improves wisdom over time, but without active maintenance the graph can degrade through duplication, contradiction, and staleness. Layer 2 performs three maintenance operations directly on the PCR structure.

**PCR-Based Deduplication.** Wisdom pairs whose $P$, $C$, and $R$ nodes are semantically equivalent (up to node normalization) are identified and merged. Evidence counts, cumulative reviews, and causal edge scores are consolidated into the surviving variant, ensuring that retrieval is not diluted by near-identical entries.

**Contradiction Resolution.** Wisdom pairs that share the same $P$ and $C$ but yield opposing $R$ outcomes—or share the same $R$ but prescribe contradictory actions under the same conditions—are detected through polarity analysis over the PCR graph. The variant with stronger operational evidence is retained; the weaker variant is either deprecated or annotated with exclusion conditions that restrict its applicability, converting a global contradiction into a context-bounded judgment.

**Evidence-Triggered Content Update.** Wisdom items that accumulate negative feedback or receive explicit update signals are candidates for automated content update. The system cross-references domain-similar wisdom within the graph and performs external validation to identify outdated facts, deprecated interfaces, or changed configurations, then applies targeted content updates while preserving the PCR structure and evidence history. This ensures that the graph reflects current operational reality rather than the state at the time of original extraction. Figure 6 summarizes the three maintenance operations.

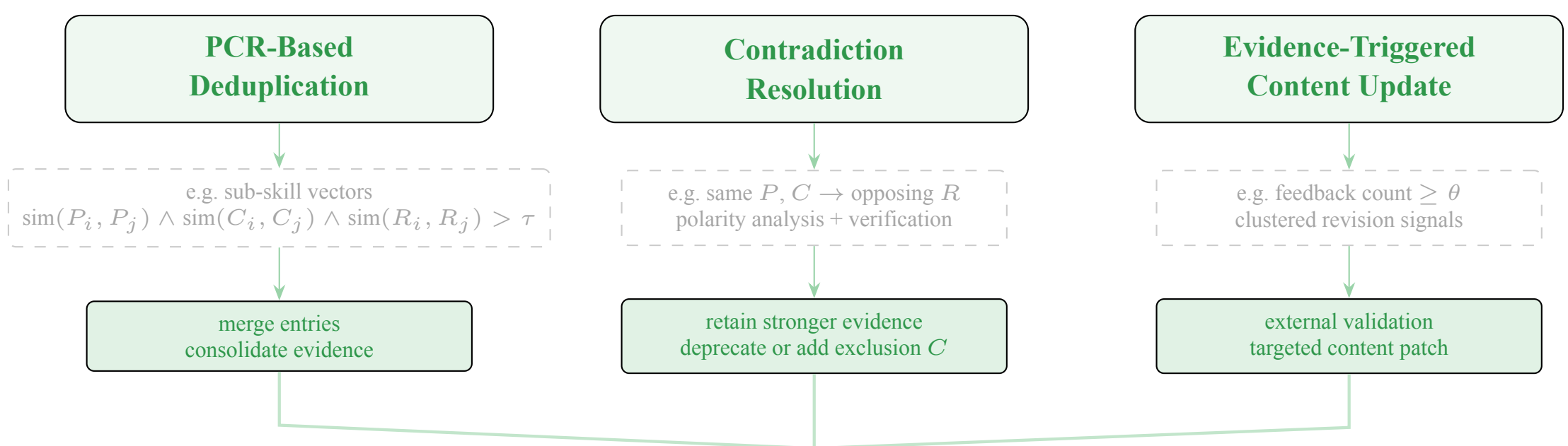

Figure 6: Three graph hygiene operations. Deduplication merges semantically equivalent PCR triplets via sub-skill vector matching. Contradiction resolution detects opposing resultants through polarity analysis and converts global conflicts into context-bounded judgments. Content update triggers on accumulated feedback and applies targeted patches while preserving evidence history.

# 6 Layer 3: Evaluation-Driven Multi-Agent Optimization

As discussed in Section 2.3, existing automation approaches share a common limitation: they do not learn optimization strategies themselves. TDD-based verification causes test infrastructure to become technical debt during workflow redesign; stateless retry loops such as Ralph Loop [11] fail to distinguish productive work from waste; and prompt/parameter optimization retains only the results while leaving "how the optimization was performed" unstructured. Layer 3 targets this gap.

There is, however, a more fundamental limitation that has received less attention. Current prompt and workflow optimizers define each LLM invocation as a module $M_i = (\pi_i, \theta_i, X_i, Y_i)$ and optimize primarily at the prompt level [13, 2]. A multi-hop QA system, for example, is decomposed into a chain of such modules—a query generator, a document summarizer, an answer extractor—each improved by refining its prompt $\pi_i$. This *prompt-centric* paradigm scales linearly in the number of modules: as task complexity grows, more modules are introduced, each duplicating preamble context in its prompt, increasing aggregate token cost and maintenance burden.

We therefore take a broader view of the optimization target. Following prior formalizations of agentic workflows [38, 6], we model an agent workflow as a directed graph $\mathcal{A} = (\mathcal{N}, \mathcal{E})$ whose nodes are of three heterogeneous types—deterministic code nodes, single LLM calls parameterized by prompt and decoding settings, and tool-using agent nodes performing multi-turn reasoning with tool invocations—and whose edges define execution order; a *workflow stage* denotes a contiguous segment of $\mathcal{A}$ corresponding to a functional role (e.g., retrieval, synthesis, verification), and serves as the granularity at which Layer 2 aligns wisdom to operational context. Under this formulation, a prompt-centric optimizer restricts optimization to the $\pi$-component of LLM-call nodes alone, whereas Layer 3 targets the full tuple (prompt, tool configuration, code logic, orchestration structure) across all node types.

Real-world agent systems, built on frameworks such as OpenAI/Claude Agent SDK, Google ADK, and Vercel AI SDK, exhibit exactly this heterogeneous structure. Not every LLM call warrants an agent node—agent nodes incur substantially higher cost due to multi-turn reasoning and tool invocation—but where a task requires tool use, context management, and conditional branching, a single agent node can subsume what would otherwise be multiple prompt-only modules. The optimization challenge is that these heterogeneous node types coexist in one workflow: a prompt change in an agent node can alter tool-calling behavior, context management strategy, and downstream control flow simultaneously. Jointly optimizing prompt, tool configuration, code logic, and orchestration structure across all node types is the scope that Layer 3 targets. Figure 7 illustrates this distinction.

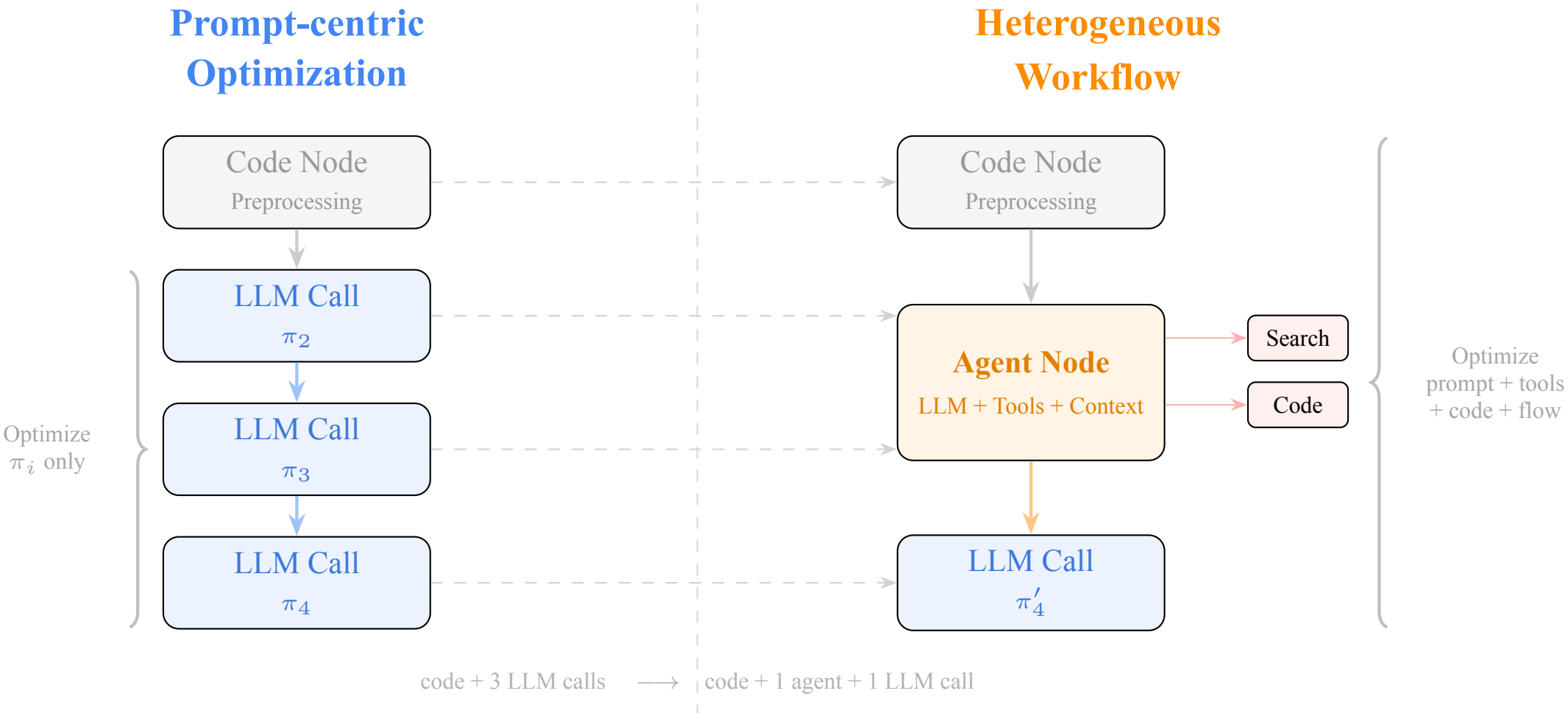

Figure 7: Prompt-centric vs. heterogeneous workflow optimization. Prompt-centric optimizers treat each LLM call as an independent module and optimize prompts $\pi_i$ individually. In a heterogeneous agent workflow, a tool-using agent node subsumes multiple LLM calls where multi-turn reasoning is required, while deterministic code nodes and simple LLM calls remain unchanged. Layer 3 jointly optimizes prompt, tool configuration, code logic, and orchestration flow across all node types.

Layer 3 addresses both the strategic learning gap and the optimization scope gap. Given a user-specified agent system and task, it operates as a multi-agent automatic optimization engine in which multiple specialized agents collaborate to iteratively improve the target system—including its agent nodes, code logic, and orchestration structure. At each cycle, a meta-learning agent extracts not only the optimization outcomes but also structured optimization trajectories capturing which strategies proved effective under which conditions, accumulating them in Layer 2. When a similar situation arises in a subsequent project, the system begins from a previously validated optimization path rather than engaging in unverified exploration. In other words, Layer 3 improves not only the agent system but also the method of optimization itself.

Where Layer 2 determines which knowledge to compose and how, Layer 3 measures whether that composition actually works and feeds the resulting evidence back to refine both the knowledge and the composition strategy.

Concretely, Layer 3 is realized as a multi-agent system in which a central optimization agent coordinates specialized sub-agents (scientist, code reviewer, redesign agent), with the Wisdom Graph providing compositional retrieval over PCR-decomposed wisdom, the Seed-Epoch regime (Section 6.4) providing attribution discipline, and the specialized-agent pipeline providing separation of concerns between diagnosis, hypothesis, execution, and measurement. The optimization gains reported in Section 7.2 therefore reflect the combined contribution of these mechanisms.

## 6.1 The Adapt Pipeline

The optimization target of Layer 3 is not a single prompt but an entire heterogeneous workflow encompassing code nodes, LLM calls, tool-using agents, orchestration logic, and evaluation pipelines. Given an existing agent system, Layer 3 inventories its assets, reverse-engineers a formal specification, generates evaluation data where absent, and initiates iterative optimization against the resulting baseline. The principle throughout is to respect the existing code structure while making targeted improvements, with mandatory user interaction gates at critical decisions—intent confirmation, data strategy approval, and architectural feedback—ensuring that the infrastructure operates as a collaborative tool rather than an unsupervised loop.

The pipeline proceeds through the following stages:

1. **Asset Scan:** Inventories the target system's source code, LLM call structure, existing evaluation met-

rics, and any available evaluation data, characterizing the current state as the optimization baseline.

2. **Adaptive Entry:** Determines the starting point of the pipeline based on what the scan surfaces—code only, PRD only, data only, or any combination thereof—skipping subsequent stages whose outputs are already present.

3. **Research Agents (conditional):** When the Asset Scan finds that an evaluation dataset or evaluation code is missing, specialized research agents are invoked in parallel to gather the external grounding required for synthesis—surveying state-of-the-art techniques and their reported performance and identifying publicly available datasets and their relevance—producing a structured report that grounds downstream decisions in external evidence. When both an evaluation dataset and evaluation code already exist, this stage is skipped.

4. **Reverse PRD / Data Synthesis:** A Reverse Engineer agent reconstructs a *Pipeline Requirement Document* (PRD) from the existing codebase, capturing the I/O schema, scenario taxonomy, evaluation criteria, and reference datasets implicit in the current system (augmented with findings from Research when invoked). When the workflow lacks an evaluation dataset—a common situation in which only a pipeline exists but no data to evaluate it against—the Data Agent generates one directly from the PRD: a data analysis step derives a category–difficulty matrix, seed scenarios spanning the full space are designed and presented at a mandatory human approval gate, and approved seeds anchor the distribution for parallel workers that produce stratified training and validation sets, together with an optional document store for retrieval-augmented scenarios. When a dataset already exists, this stage reduces to verification and hands control to the coverage-gap augmentation of Section 6.3. The generation scope is bounded by what LLM synthesis can produce—text-based input–output pairs and optional text document stores; non-textual media such as images, audio, and video fall outside the current pipeline.

5. **Shared Optimization Loop:** Executes Seed-Epoch-based iterative optimization (Section 6.4), with a central optimization agent coordinating specialized sub-agents (scientist, code reviewer, redesign agent) through a disciplined cycle of diagnosis, hypothesis, execution, and measurement.

6. **Meta-learning Agent:** At optimization completion, extracts patterns discovered during the run—which strategies worked under which conditions, which trajectories reduced failures, and which curation steps required wisdom that was not available—as structured wisdom and registers them in Layer 2; the surfaced capability gaps are then clustered and materialized as new atomic skills in the global skill library, expanding the pool that subsequent optimizations draw from.

Figure 8 illustrates the pipeline and its connection to Layer 2.

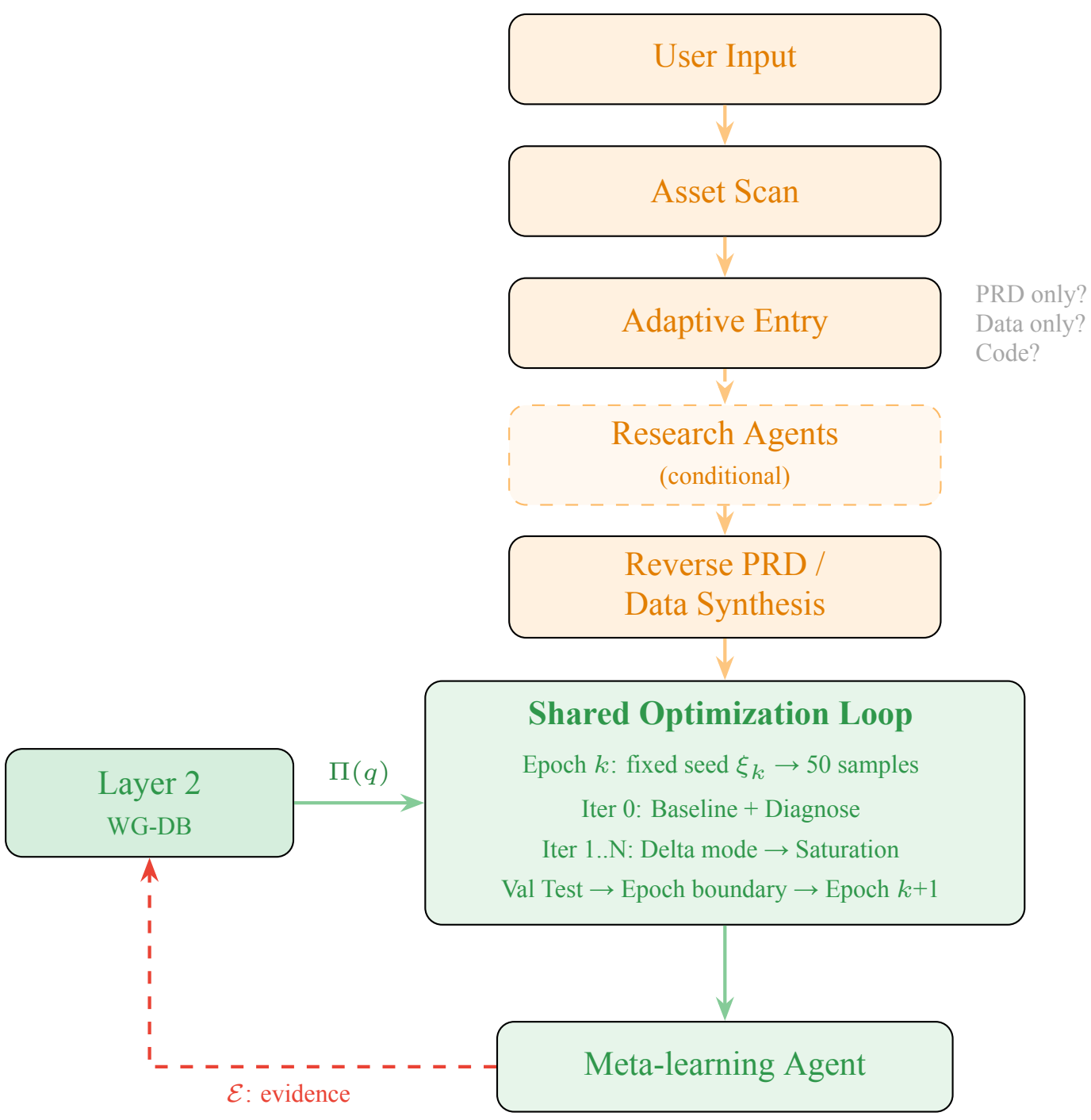

Figure 8: The Adapt pipeline of Layer 3. Asset Scan inventories existing system state; Adaptive Entry chooses the starting point and conditionally invokes Research Agents when evaluation data or evaluation code is missing; Reverse PRD reconstructs the specification and, when required, synthesizes the evaluation dataset. The pipeline then enters the Seed-Epoch-based Optimization Loop. Layer 2 (WG-DB) supplies curation plans $\Pi(q)$ to the loop, and after optimization the Meta-learning Agent feeds evidence $\mathcal{E}$ back to Layer 2, closing the cycle.

## 6.2 Multi-Agent Collaboration Inside the Optimization Loop

Layer 3 employs over a dozen specialized agents, each with strictly bounded responsibilities. Table 3 summarizes the key agents by phase.

Table 3: Some key agents in the Layer 3 pipeline.

| Phase | Agent | Role |
|---|---|---|
| Research (conditional) | SOTA Research | Surveys state-of-the-art techniques and benchmarks |
| | Dataset Search | Searches for existing open datasets |
| PRD | Reverse Engineer | Reverse-engineers existing codebase into a Pipeline Requirement Document |
| Data | Data Analysis | Designs synthetic data strategy from the PRD |
| | Data Sample Workers | 3 parallel workers by difficulty (easy/medium/hard) |
| | Data Doc Workers | 2 parallel workers for document store generation |
| | Data Downloader | Downloads and prepares open datasets |
| Optimization | Scientist | Error pattern analysis with root-cause vs. victim distinction and ROI-based prioritization |
| | Code Reviewer | Static analysis + dry-run validation + anti-heuristic enforcement |
| | Optimization Agent | Central deliberation: diagnosis → hypothesis → fix → measurement |
| | Redesign Agent | Architectural restructuring when incremental optimization stagnates |
| Post-opt. | Meta-learning Agent | Extracts optimization trajectories and registers them in Layer 2 |

The optimization loop proceeds in two stages. In **Iteration 0 (Baseline Analysis)**, the initial system is evaluated and the scientist agent performs multi-faceted diagnostics: error pattern classification by fixability (prompt-fixable, code-fixable, architecture-required), per-node root-cause analysis that distinguishes genuine failure sources from downstream victims via error propagation tracing, and ROI-based prioritization that ranks nodes by (error rate $\times$ case count) / cost. A five-signal architectural verdict—error concentration, context loss ratio, error type diversity, uniform failure rate across LLM nodes, and user goal gap—determines whether incremental improvement or full architectural redesign is warranted; three or more signals trigger redesign, with the goal gap signal carrying sufficient weight to trigger redesign on its own. The system then computes a target accuracy based on the fixability distribution of observed errors, rather than relying on arbitrary targets.

**Iteration 1+ (Delta Mode):** The central optimization agent directly performs delta analysis and repeats fixed-seed evaluation under the Seed-Epoch structure. At each iteration, the agent applies up to three targeted fixes (prompt improvement, structural code fix, or parameter tuning), submits them to the code reviewer for validation, and measures the result. The code reviewer performs static analysis and dry-run testing before each evaluation, catching runtime errors and spec violations before they consume evaluation budget.

The architectural principle is centralized deliberation with specialized execution. In deep delegation chains, context is lost at each boundary and optimization direction becomes diluted. Layer 3 avoids this: the central optimization agent directly decides "what is the problem, what should be tried, and how should the results be interpreted," while specialized agents perform only well-bounded tasks. From the user's perspective, optimization proceeds as a systematic cycle of diagnosis $\rightarrow$ hypothesis $\rightarrow$ execution $\rightarrow$ measurement $\rightarrow$ next hypothesis.

## 6.3 Data-Aware Optimization

Most agent optimization systems treat training data as fixed and optimize only the workflow [13, 21, 2]. Layer 3 extends the optimization scope to include the data itself: before entering the workflow optimization loop, the system analyzes whether observed failures stem from workflow limitations or from gaps in training data coverage, and intervenes accordingly.

Layer 3 addresses this through a conditional data augmentation step that executes between baseline evaluation and the optimization loop. The system first classifies the evaluation structure into one of three types: discrete checkers with enumerable scoring units, rubric-based evaluation with natural-language criteria, or continuous metrics without explicit dimensions. For each type, a type-specific method discovers the full set of evaluation dimensions—by static analysis of evaluation code for discrete checkers, by extracting axes from rubric prompts, or by clustering inputs and correlating with metric scores for continuous cases.

Each discovered dimension is then assessed along three axes: density (how many training examples cover it), diversity (how varied those examples are), and baseline performance. The critical insight is the distinction between *data gaps* and *workflow problems*: a dimension with sufficient, diverse training data but low performance indicates a workflow problem that augmentation cannot solve, while a dimension with absent or sparse data indicates a genuine coverage gap. Augmentation targets only the latter.

When gaps are identified, targeted synthetic data is generated for the specific missing dimensions, merged with the original training set under a synthetic ratio bounded by the gap-detection signal itself—scaled to the magnitude of the detected coverage deficit so that augmentation closes the gap without overwhelming the empirical distribution, and validated through dry-run evaluation. The baseline is then re-evaluated on the augmented dataset before the optimization loop begins. When no gaps are found, this step is skipped entirely with zero additional latency. Figure 9 illustrates this conditional flow.

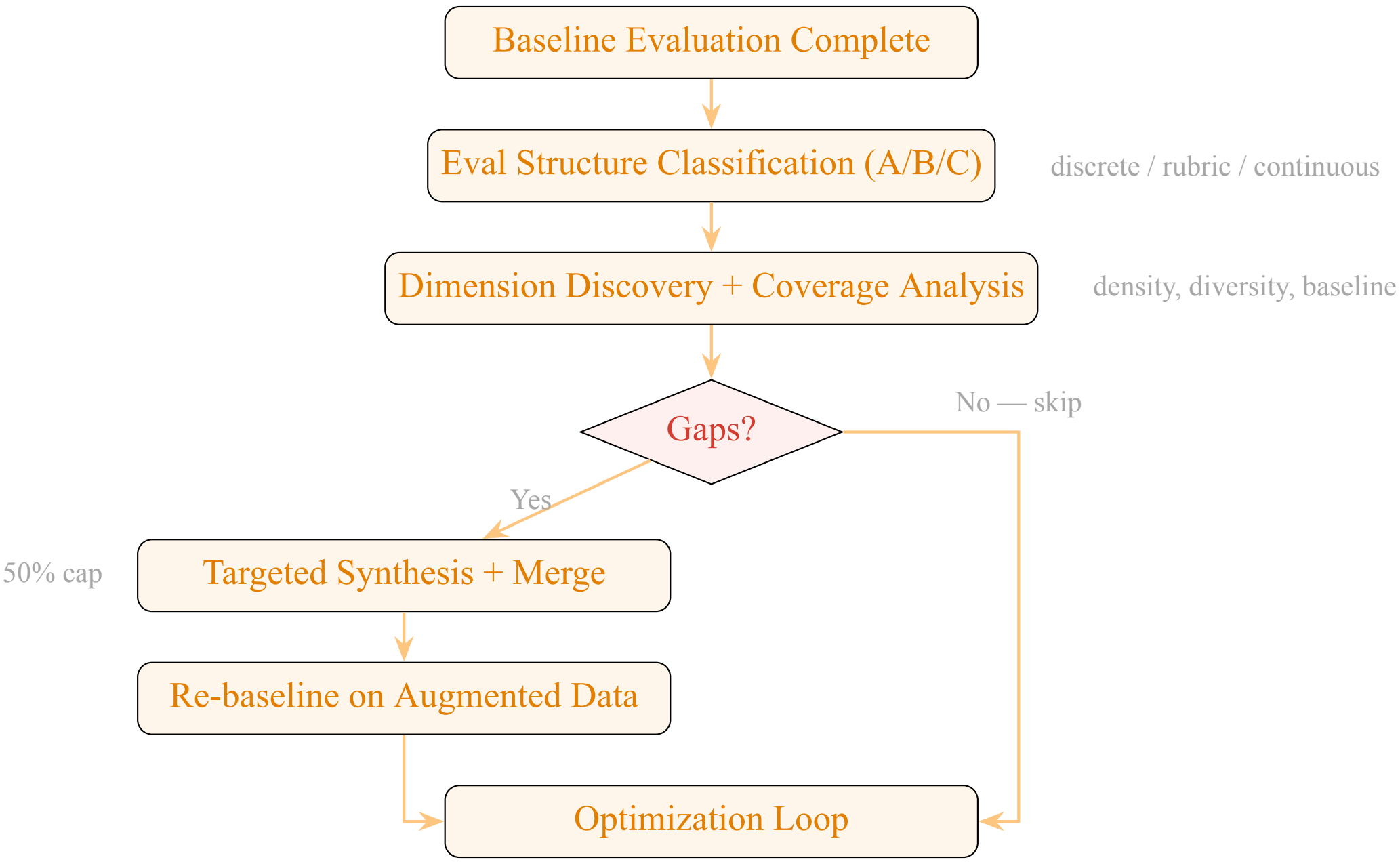

Figure 9: Coverage-driven data augmentation. The system classifies the evaluation structure, discovers evaluation dimensions, and distinguishes data gaps from workflow problems. Augmentation executes only when genuine coverage gaps exist; otherwise, the step is skipped with zero overhead.

### 6.4 Seed-Epoch: Overfitting Regulation

Even when multiple agents iteratively improve a system, whether those improvements are genuine is a separate question. If evaluation data is resampled at every iteration, it becomes impossible to determine whether observed performance gains result from strategy changes or simply from drawing easier samples. This "meaningless delta" problem structurally undermines the reliability of optimization.

**Definition 2** (Seed-Epoch)**.** An evaluation dataset $D_e = \text{Sample}(\mathcal{D}_{\text{train}}, n, \xi_e)$ fixed by a deterministic seed $\xi_e$ at epoch $e$. All iterations $i$ within an epoch are evaluated on the same $D_e$:

$$\Delta_i = \text{Acc}(D_e, v_i) - \text{Acc}(D_e, v_{i-1}) \tag{11}$$

Since the evaluation dataset is identical for all iterations within the same epoch $e$, data variance is eliminated, and $\Delta_i$ can attribute its primary source of variation to the effect of strategy changes. Here, $v_i$ denotes the $i$-th system variant.

The core idea is "changing only the strategy while keeping the same test problems." This discipline ensures that instead of merely an impression that performance improved, there is a record of which changes produced which differences under identical conditions. Layer 2 uses this attributed evidence to decide which wisdom to promote and which to defer. If attribution is infeasible, the graph becomes contaminated with an indiscriminate mixture of success stories and coincidences.

When $|\Delta_i| < \epsilon$ for $m$ consecutive iterations, optimization within the current epoch is deemed saturated, and the epoch advances to a new seed $\xi_{e+1}$. The new epoch evaluates on fresh data, thereby naturally validating whether strategies overfitted to the previous epoch.

A complementary design choice governs *when* to run validation. Running validation at every iteration is wasteful; running it only at epoch boundaries risks missing critical generalization signals mid-epoch. MEGA adopts a momentum-triggered validation policy: validation is executed at three points—(1) the baseline iteration, (2) epoch boundaries, and (3) momentum events, defined as recovery from a regression or the achievement of a new best score. This selective scheduling concentrates validation budget on iterations where generalization information is most actionable.

### 6.5 Optimization Safeguards

Seed-Epoch ensures that performance deltas are attributable. Three additional safeguards ensure that the improvements themselves are genuine.

**Anti-Heuristic Enforcement.** Code modifications are restricted to structural transforms: JSON parsing, encoding normalization, null handling, type coercion, and error recovery. Content-aware patterns—regular expressions that match specific answer formats, conditional branches for particular question types, and lookup tables mapping inputs to outputs—are prohibited. The code reviewer agent enforces this constraint through static analysis and dry-run testing before each evaluation, rejecting fixes that would overfit to the training set rather than improve the system's general capability.

**Overfitting Detection.** The system tracks the gap between train and validation accuracy across epochs. When the gap exceeds a threshold at epoch saturation, the next epoch restricts optimization to prompt improvements only, prohibiting code fixes that are more susceptible to training-set-specific patterns. This constraint, combined with Seed-Epoch's automatic generalization check upon epoch transition, provides a layered defense against overfitting.

**Fixability-Based Goal Calibration.** Rather than pursuing an arbitrary accuracy target, the system computes a realistic ceiling from the baseline error distribution. Errors classified as prompt-fixable receive a high recovery estimate; code-fixable errors receive a moderate estimate; architecture-required errors define a hard ceiling. The resulting target reflects what the current system structure can realistically achieve, preventing wasted iterations on unreachable goals.

### 6.6 Cross-Layer Compounding Cycle

Most optimization systems improve performance on the current project and stop. When the same type of problem is encountered in a subsequent project, there is no structural mechanism to reuse insights discovered during previous optimization. Layer 3 closes this gap by feeding structured evidence back to Layer 2 after each optimization cycle.

**Definition 3** (Verdict)**.** A *verdict* is a multi-dimensional judgment tuple $(\Delta_{\text{acc}}, \Delta_{\text{lat}}, \Delta_{\text{tok}})$ recording the accuracy delta, latency change, and token usage change observed after applying a wisdom-informed strategy change under fixed-seed evaluation.

Every verdict generated during optimization is fed back to Layer 2 to update two quantities:

- $\eta(\omega)$: The evidence confidence of wisdom that received an effective verdict is increased, boosting it in subsequent retrieval, while repeatedly failed wisdom is pushed to lower ranks.
- $\tau(\cdot)$: The historical transfer rate for the corresponding similarity range and workflow stage is updated with the observed outcome, making initial $p$ estimates for similar new wisdom more accurate in subsequent cycles.

These verdicts are not single scalars. If a strategy improved accuracy but doubled latency, it becomes conditional wisdom valid only in certain contexts. MEGA does not treat "good wisdom" as an absolute category; instead, it accumulates wisdom that has been validated conditionally on context.

Furthermore, the meta-learning agent extracts patterns discovered during the optimization process itself—which strategy combinations worked under which conditions, which trajectories reduced failures—as structured optimization trajectories and registers them in Layer 2. These trajectories enable the system, when encountering a similar situation in a future project, to start from a validated path rather than exploring from scratch.

This is what makes Layer 3 not a one-shot optimizer but a self-reinforcing optimization engine that drives the compounding cycle of the entire MEGA framework. Within Layer 3, the cycle of diagnosis $\rightarrow$ hypothesis $\rightarrow$ execution $\rightarrow$ measurement repeats in a self-reinforcing manner, and its outputs refine Layer 2 so that the next project starts from a higher baseline.

# 7 Experiments and Evaluation

The evaluation of MEGA is organized along two axes: (1) whether the curation quality of accumulated wisdom translates into improved downstream performance, and (2) whether MEGA's optimization surpasses existing LM-centric optimizers. Experiment configurations, optimization, and per-benchmark results for both axes are available at `https://github.com/mega-edo/mega_benchmark`.

## 7.1 Wisdom Curation Quality

To measure whether WG-DB's compositional curation translates into downstream task performance, we evaluate on SkillsBench [17], a benchmark of 84 tasks across 11 domains where each task is verified by deterministic pytest assertions in isolated Docker containers with binary pass/fail verdicts.

**Setup.** All conditions run on a GCP `n2d-standard-128` VM using Gemini CLI with Gemini 3 Flash as the agent, with 5 attempts per task and 43 concurrent Docker trials. The skill pool comprises 4,207 assets including SkillsBench's golden skills for all curation systems. Gemini 3 Flash is also used for all LLM components in the curation pipelines so that no condition benefits from a stronger model at any stage. We compare four conditions that share the same agent, skill pool, tasks, verifiers, and Docker environment; the only variable is the skill discovery and orchestration method:

- **No Skills**: the agent receives only the task instruction.
- **SkillNet** [16]: LLM-based metadata matching followed by LLM-generated procedural guidance. We use SkillNet's original prompts (`retrieve_relevant_skills_prompt`, `generate_overall_procedure_prompt`) without modification.[1]
- **AgentSkillOS** [15]: hierarchical tree search via the original `TreeBuilder` and `Searcher` modules, followed by Quality-First DAG plan generation via the original `build_planner_prompt`—the best-performing strategy reported by the authors.[2] The DAG is topologically linearized for SkillsBench's sequential execution environment.
- **MEGA (WG)**: PCST-based compositional retrieval from WG-DB, followed by role-differentiated plan assembly (Section 5).

Each method retrieves from the shared pool and the retrieved subset is injected into execution. The reported results therefore reflect end-to-end pipeline performance in which retrieval and curation are measured jointly, meaning an inaccurate curation plan can degrade performance even when retrieval is accurate. In deployment, where all available skills' summaries are already loaded into the agent's context, curation quality becomes the dominant factor; the benchmark setting, by contrast, focuses on the retrieval–curation pipeline as a whole.

**Results.** Table 4 reports pass rate, average token consumption per task, curation latency (time to discover and orchestrate skills before execution), and curation ROI—defined as pass-rate improvement per additional megatoken consumed relative to the No Skills baseline.

Table 4: Wisdom curation quality on SkillsBench (84 tasks, Gemini 3 Flash). Pass rate is the primary metric; token usage and curation latency capture efficiency. Efficiency is defined as pass rate per megatoken consumed (score/Mtok).

| | **Pass Rate** (%) | **Avg Tokens/Task** (k) | **Curation Latency** (sec/task) | **Efficiency** (score/Mtok) |
|---|---|---|---|---|
| No Skills | 31.5 | 894 | — | 0.353 |
| AgentSkillOS | 41.1 | 1189 | 403.4 | 0.345 |
| SkillNet | 41.7 | 983 | 37.8 | 0.424 |
| MEGA (WG) | **46.5** | **822** | **11.8** | **0.566** |

[1] `https://github.com/zjunlp/SkillNet/tree/main/experiments`
[2] `https://github.com/ynulihao/AgentSkillOS/tree/main/src`

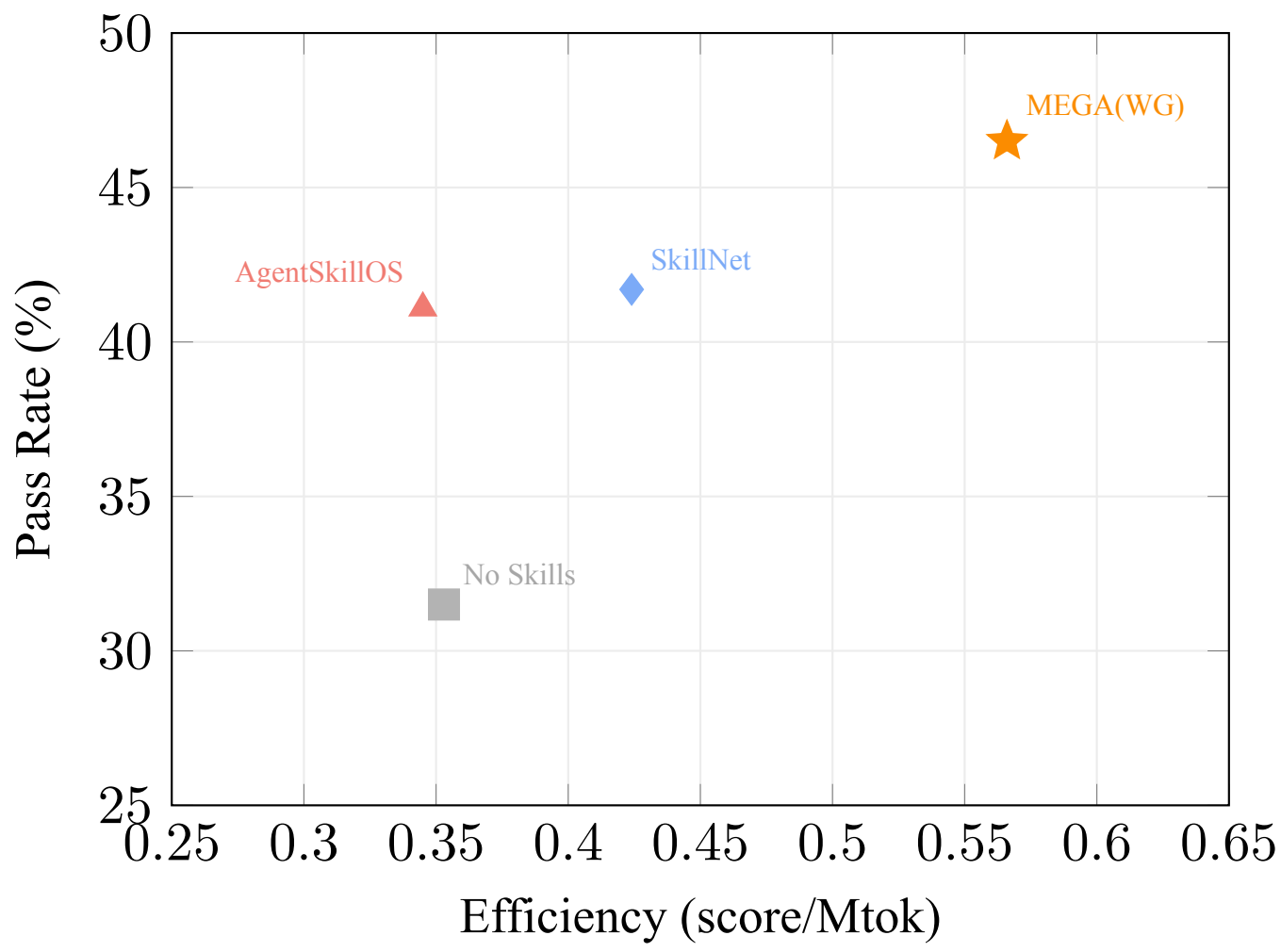

Figure 10: Pass rate vs. efficiency on SkillsBench. Higher and further right is better.

Three observations stand out. First, skill curation is not automatically beneficial—the method of curation matters. AgentSkillOS achieves 41.1% and SkillNet reaches 41.7%, both substantially above the No Skills baseline (31.5%), confirming that skill curation provides meaningful lift. However, AgentSkillOS incurs the highest token cost (1189k) and curation latency (403.4s) for a pass rate comparable to SkillNet's simpler two-call approach (983k tokens, 37.8s). MEGA (WG) achieves the highest pass rate (46.5%) with the best efficiency (0.566 score/Mtok), indicating that *how* skills are selected and composed—not merely *that* they are provided—determines downstream performance. Second, token consumption reveals a key difference. Both SkillNet and AgentSkillOS inflate context by injecting whole-skill content (procedural text or linearized DAG with per-step skill references), while MEGA (WG) injects only the atomic PCR units relevant to the task through role-differentiated placement, reducing context overhead while preserving actionable guidance.

Third, curation latency varies dramatically across methods due to differing LLM dependencies. AgentSkillOS requires sequential LLM calls at each level of its capability tree during skill discovery (up to six levels), plus an additional call for DAG plan generation, resulting in the highest per-task latency. SkillNet requires two LLM calls per task (skill retrieval and procedure generation). MEGA (WG)'s PCST retrieval operates over a pre-indexed graph with subgraph pruning and requires no LLM calls during the retrieval phase itself, yielding the lowest curation latency. Together, these results indicate that compositional retrieval over a typed Wisdom Graph—where skills are decomposed into atomic units and assembled through graph reasoning—outperforms both flat metadata matching and hierarchical tree search, not only in task performance but also in token efficiency and curation speed.

## 7.2 Optimization Performance

To measure whether MEGA's optimization surpasses existing LM-centric optimizers, we compare against MIPROv2 [21], TextGrad [37], GEPA [2], and Feedback Descent [14] on GPT-4.1 Mini. All baselines operate on ground-truth evaluation data, and MEGA is run against the same ground-truth datasets for a fair comparison.

**Setup.** Each benchmark is paired with a compound AI system (multi-hop retriever, instruction-following rewriter, claim verifier, or privacy-aware delegator) taken from GEPA's released evaluation code.[3] We strip all DSPy and GEPA optimizer dependencies to obtain a pure workflow—i.e., a bare multi-stage LLM pipeline with no optimizer-specific scaffolding—and use this as both the starting point for MEGA's optimization and the baseline row in Table 5. The evaluation checkers and scoring artifacts remain identical to those in the GEPA codebase, ensuring that all methods are measured under the same grading criteria. Baseline scores for MIPROv2, TextGrad, and GEPA are cited directly from [2]; Feedback Descent scores are cited from [14]. The evaluation uses four benchmarks from GEPA's suite with identical train/test splits and seeds: HotpotQA (150/100/300), IFBench (150/100/294), HoVer (150/100/300), and PUPA (111/111/221), where each tuple denotes (train/val/test). MEGA uses smaller validation sets than GEPA's original splits (e.g., 100 vs. 300 for HotpotQA) to reduce optimization cost. These benchmarks collectively require diverse capabilities including multi-hop reasoning, instruction following, retrieval-based verification, and privacy-preserving delegation. MEGA's Layer 3 optimization runs as a multi-agent loop, drawing from the same 4,207-asset Wisdom Graph evaluated in Section 7.1.

[3] `https://github.com/gepa-ai/gepa-artifact/tree/main/gepa_artifact/benchmarks`

Table 5: Benchmark results on GPT-4.1 Mini (`gpt-4.1-mini-2025-04-14`). Baseline, MIPROv2, TextGrad, and GEPA scores are cited from GEPA [2]. Feedback Descent scores are cited from [14]. GEPA (best) reports the per-benchmark maximum of GEPA and GEPA+Merge variants.

| | **HotpotQA** | **IFBench** | **HoVer** | **PUPA** | **Agg.** |
|---|---|---|---|---|---|
| Baseline | 38.00 | 47.79 | 46.33 | 78.57 | 52.67 |
| MIPROv2 | 58.00 | 49.15 | 48.33 | 83.37 | 59.71 |
| TextGrad | 62.33 | 48.64 | 47.67 | 85.68 | 61.08 |
| Feedback Descent | 68.33 | 54.59 | 57.67 | 85.66 | 66.56 |
| GEPA (best) | 69.00 | 55.95 | 56.67 | 96.46 | 69.52 |
| MEGA | **72.67** | **61.05** | **74.67** | **97.81** | **76.55** |

**Results.** MEGA achieves an aggregate score of 76.55, outperforming the strongest baseline GEPA (69.52) by +7.03 points and the next-strongest Feedback Descent (66.56) by +9.99 points. The improvement is consistent across all four benchmarks. The largest absolute gain appears on HoVer (+18.00 over GEPA), where multi-hop retrieval benefits most from compositional wisdom that chains retrieval strategies across hops. On IFBench, MEGA achieves +5.10 over GEPA, suggesting that structured constraint-handling wisdom transfers effectively to instruction-following tasks. HotpotQA (+3.67) and PUPA (+1.35) show consistent gains; both benchmarks approach the ceiling reachable through workflow and prompt optimization on GPT-4.1 Mini, beyond which further improvement requires a more capable base model. Notably, these results are achieved with smaller validation sets than GEPA's original splits, indicating that MEGA's optimization loop requires less validation data to converge.

Within Layer 3, the Wisdom Graph supplies compositional retrieval over PCR-decomposed wisdom, and the Seed-Epoch regime supplies the attribution discipline that isolates strategy-driven gains from data variance. Since the baseline optimizers themselves operate over capable LLMs, the measured gap reflects what these structural elements add on top of capable agentic reasoning. The above gains therefore measure the integrated infrastructure—a curated Wisdom Graph supplying compositional context, plus evaluation-driven optimization with disciplined attribution—rather than any single component.

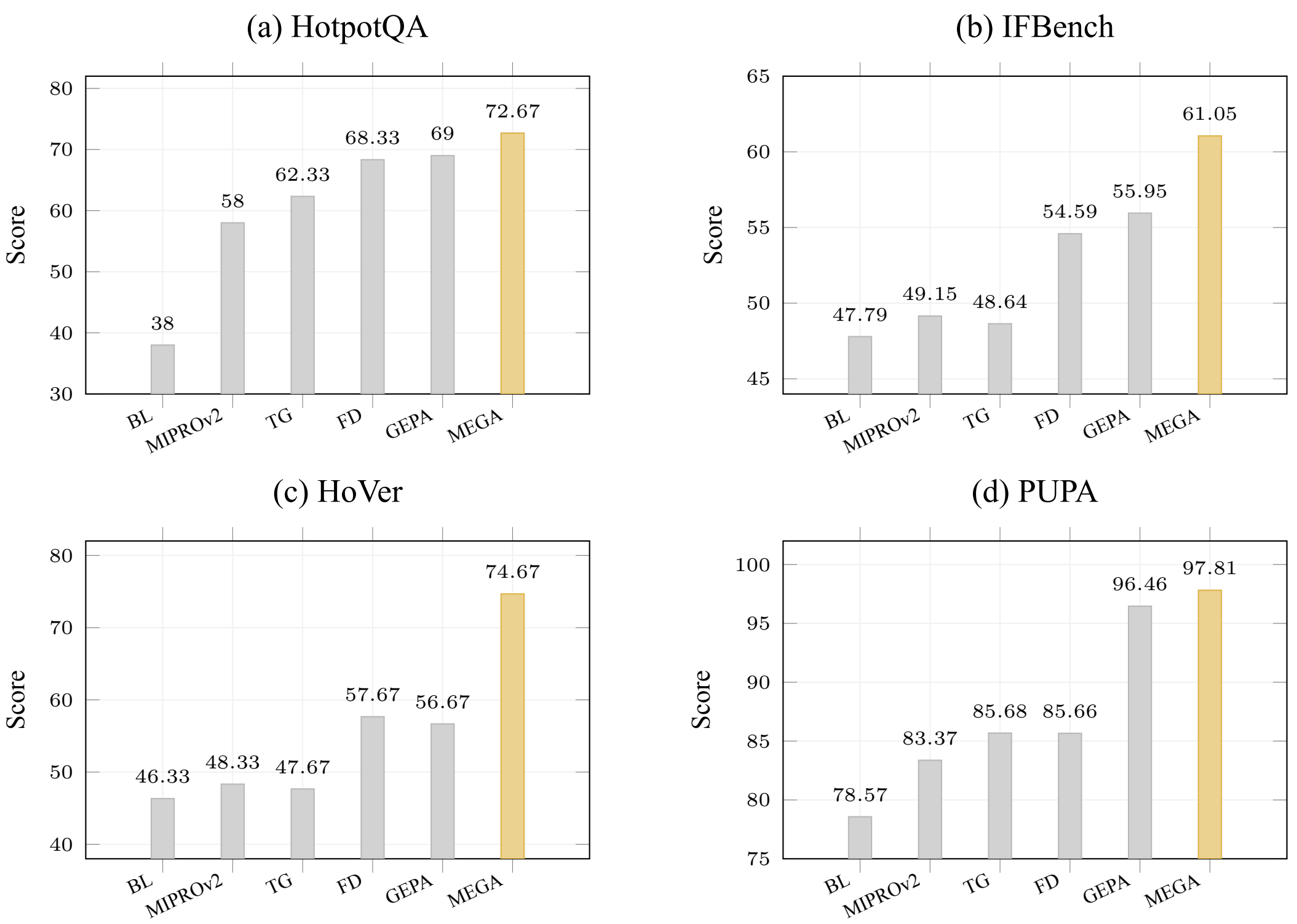

Figure 11: Per-benchmark comparison on GPT-4.1 Mini across four benchmarks. Each panel uses a benchmark-specific y-axis range. Values are shown above each bar. BL = Baseline, TG = TextGrad, FD = Feedback Descent.

# 8 Conclusion

This report presents MEGA, an infrastructure that unifies evaluation-driven agent optimization with self-evolving wisdom curation. Layer 1 distills agent sessions into verified knowledge through behavioral A/B validation, ensuring that only empirically confirmed wisdom enters the system. Layer 2 decomposes this knowledge into atomic PCR units within a typed Wisdom Graph, performs logical reasoning to discover implicit compositional relations, and assembles context-specific execution plans through compositional retrieval. Layer 3 jointly optimizes heterogeneous agent workflows—code nodes, LLM calls, and tool-using agents—and attributes performance changes to specific strategy modifications, producing operational evidence that continuously refines both the Wisdom Graph and the curation logic that governs it.

Empirical results support this design along two axes. On SkillsBench (84 tasks, 11 domains), MEGA achieves a 46.5% pass rate with an efficiency of 0.566 score/Mtok, surpassing SkillNet (41.7%, 0.424) and AgentSkillOS (41.1%, 0.345) while consuming fewer tokens per task. On four established workflow benchmarks, MEGA attains an aggregate score of 76.55 on GPT-4.1 Mini—7.03 points above the strongest reported baseline. These gains reflect the architecture as a whole: behavioral validation gating graph entries, compositional reasoning shaping execution plans, and attributable measurement isolating strategy-driven effects.

These results are not attributable to any single layer in isolation. Layer 1's behavioral validation controls the precision of graph entries; Layer 2's compositional reasoning determines the structural quality of execution plans; Layer 3's attributable measurement provides the empirical signal that calibrates both. The three layers form a closed loop—distillation feeds reasoning, reasoning guides optimization, and optimization produces evidence that refines both the knowledge and the curation logic—and the observed gains reflect this loop operating end to end.

The deeper architectural significance lies in this feedback cycle. Each optimization run refines not only the target system but also the curation strategies and optimization trajectories within the Wisdom Graph—the optimizer optimizes itself. As this cycle repeats across projects, the graph is intended to accumulate validated compositional patterns so that new projects begin from a baseline shaped by prior optimization rather than from scratch. In this design, optimizing an agent system and optimizing the very process by which optimization is performed are one and the same—a meta-optimization that compounds over time.

## References

[1] Russell L. Ackoff. From data to wisdom. *Journal of Applied Systems Analysis*, 16:3–9, 1989.

[2] Lakshya A Agrawal, Shangyin Tan, Dilara Soylu, Noah Ziems, Rishi Khare, Krista Opsahl-Ong, Arnav Singhvi, et al. GEPA: Reflective prompt evolution can outperform reinforcement learning. In *International Conference on Learning Representations (ICLR)*, 2026. Oral Presentation.

[3] Zhuoqun Chen et al. Agent KB: Leveraging cross-domain experience for agentic problem solving, 2025. arXiv:2507.06229.

[4] Darren Edge, Ha Trinh, Newman Cheng, Joshua Bradley, Alex Chao, Apurva Mody, Steven Truitt, and Jonathan Larson. From local to global: A graph RAG approach to query-focused summarization. *arXiv preprint arXiv:2404.16130*, 2024.

[5] Michel X Goemans and David P Williamson. A general approximation technique for constrained forest problems. *SIAM Journal on Computing*, 24(2):296–317, 1995.

[6] Boye Gu et al. Flow: Modularized agentic workflow automation. In *International Conference on Learning Representations (ICLR)*, 2025.

[7] Siyuan Guo et al. DS-Agent: Automated data science by empowering large language models with case-based reasoning. In *International Conference on Machine Learning (ICML)*, 2024.

[8] Zirui Guo et al. LightRAG: Simple and fast retrieval-augmented generation, 2024. arXiv:2410.05779.

[9] Bernal Jiménez Gutierrez et al. From RAG to memory: Non-parametric continual learning for large language models. In *International Conference on Machine Learning (ICML)*, 2025.

[10] Hugging Face. Upskill: Agent skill generation and evaluation. `https://github.com/huggingface/upskill`, 2025.

[11] Geoffrey Huntley. Everything is a ralph loop. `https://ghuntley.com/loop/`, 2025. Autonomous agent loop pattern for iterative task completion.

[12] Thomas Joshi, Shayan Chowdhury, and Faruk Uysal. SWE-Bench-CL: Continual learning for coding agents, 2025. arXiv:2507.00014.

[13] Omar Khattab, Arnav Singhvi, Paridhi Maheshwari, Zhiyuan Zhang, Keshav Santhanam, Sri Vardhamanan, Saiful Haq, Ashutosh Sharma, Thomas T. Joshi, Hanna Mober, et al. DSPy: Compiling declarative language model calls into state-of-the-art pipelines. In *International Conference on Learning Representations (ICLR)*, 2024.

[14] Yoonho Lee, Joseph Boen, and Chelsea Finn. Feedback descent: Open-ended text optimization via pairwise comparison. *arXiv preprint arXiv:2511.07919*, 2025.

[15] Hao Li, Chunjiang Mu, Jianhao Chen, Siyue Ren, Zhiyao Cui, Yiqun Zhang, Lei Bai, and Shuyue Hu. Organizing, orchestrating, and benchmarking agent skills at ecosystem scale. *arXiv preprint arXiv:2603.02176*, 2026.

[16] Hao Li et al. SkillNet: Create, evaluate, and connect AI skills. *arXiv preprint arXiv:2603.04448*, 2026.

[17] Xiangyi Li, Wenbo Chen, Yimin Liu, et al. SkillsBench: Benchmarking how well agent skills work across diverse tasks, 2026. arXiv:2602.12670.

[18] Jiarong Liang, Zhiheng Lyu, Zijie Liu, Xiangchao Chen, Ping Nie, Kai Zou, and Wenhu Chen. SWE-Next: Scalable real-world software engineering tasks for agents. *arXiv preprint arXiv:2603.20691*, 2026.

[19] Aman Madaan, Niket Tandon, Prakhar Gupta, Skyler Hallinan, Luyu Gao, Sarah Wiegreffe, Uri Alon, Nouha Dziri, Shrimai Prabhumoye, Yiming Yang, et al. Self-refine: Iterative refinement with self-feedback. In *Advances in Neural Information Processing Systems (NeurIPS)*, 2023.

[20] Haoran Mi et al. ProcMEM: Learning reusable procedural memory from experience via non-parametric PPO for LLM agents, 2026. arXiv:2602.01869.

[21] Krista Opsahl-Ong, Michael J. Ryan, Josh Hardy, Shwetha Patel, Michael S. Bernstein, and Christopher Potts. Optimizing instructions and demonstrations for multi-stage language model programs. In *Empirical Methods in Natural Language Processing (EMNLP)*, 2024.

[22] Siru Ouyang, Jun Yan, I-Hung Hsu, Yanfei Chen, Ke Jiang, Zifeng Wang, Rujun Han, Long T. Le, et al. ReasoningBank: Scaling agent self-evolving with reasoning memory. In *International Conference on Learning Representations (ICLR)*, 2026. arXiv:2509.25140.

[23] Charles Packer, Sarah Wooders, Kevin Lin, Vivian Fang, Shishir G. Patil, Ion Stoica, and Joseph E. Gonzalez. MemGPT: Towards LLMs as operating systems, 2023. arXiv:2310.08560.

[24] Shrey Pandit, Xuan-Phi Nguyen, Yifei Ming, Austin Xu, Jiayu Wang, Caiming Xiong, and Shafiq Joty. Synthesizing agentic data for web agents with progressive difficulty enhancement mechanisms. *arXiv preprint arXiv:2510.13913*, 2025.

[25] Judea Pearl. *Causality: Models, Reasoning, and Inference*. Cambridge University Press, 2nd edition, 2009.

[26] Ofir Press, Muru Zhang, Sewon Min, Ludwig Schmidt, Noah A Smith, and Mike Lewis. Measuring and narrowing the compositionality gap in language models. In *Findings of the Association for Computational Linguistics (EMNLP)*, 2023.

[27] Maarten Sap, Ronan Le Bras, Emily Allaway, Chandra Bhagavatula, Nicholas Lourie, Hannah Rashkin, Brendan Roof, Noah A Smith, and Yejin Choi. ATOMIC: An atlas of machine commonsense for if-then reasoning. In *AAAI Conference on Artificial Intelligence*, 2019.

[28] Noah Shinn, Federico Cassano, Ashwin Gopinath, Karthik Narasimhan, and Shunyu Yao. Reflexion: Language agents with verbal reinforcement learning. In *Advances in Neural Information Processing Systems (NeurIPS)*, 2023.

[29] Robyn Speer, Joshua Chin, and Catherine Havasi. ConceptNet 5.5: An open multilingual graph of general knowledge. In *AAAI Conference on Artificial Intelligence*, 2017.

[30] Mirac Suzgun et al. Dynamic cheatsheet: Adaptive memory for test-time learning. *arXiv preprint*, 2025.

[31] Guanzhi Wang, Yuqi Xie, Yunfan Jiang, Ajay Mandlekar, Chaowei Xiao, Yuke Zhu, Linxi Fan, and Anima Anandkumar. Voyager: An open-ended embodied agent with large language models. *arXiv preprint arXiv:2305.16291*, 2023.

[32] Jun Wang. Stateful reflective developer platforms (SRDP): Ai-native platforms for compounding agent productivity. *arXiv preprint arXiv:2512.22716*, 2025. Defines the SRDP framework on which Memento-Skills [41] is built.

[33] Zora Zhiruo Wang, Jiayuan Mao, Daniel Fried, and Graham Neubig. Agent workflow memory. *arXiv preprint arXiv:2409.07429*, 2024.

[34] Yuxin Wu et al. Optimas: Optimizing compound AI systems with globally aligned local rewards. *arXiv preprint*, 2025.

[35] Xingyao Xiang et al. SWE-Exp: Experience-driven software issue resolution, 2025. arXiv:2507.23361.

[36] Boxi Yu, Yuxuan Zhu, Pinjia He, and Daniel Kang. UTBoost: Rigorous evaluation of coding agents on SWE-Bench. In *Association for Computational Linguistics (ACL)*, 2025. arXiv:2506.09289.

[37] Mert Yuksekgonul, Federico Bianchi, Joseph Boen, Sheng Liu, Zhi Huang, Carlos Guestrin, and James Zou. TextGrad: Automatic differentiation via text. *Nature*, 2025.

[38] Jiayi Zhang et al. AFlow: Automating agentic workflow generation. In *International Conference on Learning Representations (ICLR)*, 2025. Oral Presentation.

[39] Qizheng Zhang, Changran Hu, et al. Agentic context engineering: Evolving contexts for self-improving language models. In *International Conference on Learning Representations (ICLR)*, 2026. arXiv:2510.04618.

[40] Tian Zhang, Raghu Ramakrishnan, and Miron Livny. BIRCH: An efficient data clustering method for very large databases. In *ACM SIGMOD International Conference on Management of Data*, 1996.

[41] Huichi Zhou, Siyuan Guo, Anjie Liu, Zhongwei Yu, Ziqin Gong, Bowen Zhao, et al. Memento-skills: Let agents design agents. *arXiv preprint arXiv:2603.18743*, 2026.

[42] Yuhang Zhou, Lizhu Zhang, Yifan Wu, Jiayi Liu, Xiangjun Fan, Zhuokai Zhao, and Hong Yan. Synthetic sandbox for training machine learning engineering agents. *arXiv preprint arXiv:2604.04872*, 2026.